\documentclass[10pt,twocolumn,letterpaper]{article}

\usepackage[pagenumbers]{cvpr} 

\usepackage{algorithm}
\usepackage[noend]{algpseudocode}
\usepackage{booktabs,multirow}
\usepackage[table]{xcolor}
\usepackage{listings}
\usepackage[table]{xcolor}
\definecolor{AcadGreen}{HTML}{2E7D32} 
\definecolor{AcadRed}{HTML}{C62828}   
\definecolor{AcadBlue}{HTML}{1565C0} 

\definecolor{cvprblue}{rgb}{0.21,0.49,0.74}
\usepackage[pagebackref,breaklinks,colorlinks,allcolors=cvprblue]{hyperref}
\usepackage{mdframed}   
\usepackage{fvextra}
\usepackage{marvosym}
\usepackage{fancyvrb}
\def\paperID{5212} 
\def\confName{CVPR}
\def\confYear{2026}

\title{Action-Sketcher: From Reasoning to Action via Visual Sketches for Long-Horizon Robotic Manipulation}

\author{\vspace{0.25em} Huajie Tan$^{1,2,*,\dagger}$, Peterson Co$^{1,2,*}$, Yijie Xu$^{2,3,*}$, Shanyu Rong$^{1,2}$,  Yuheng Ji$^{2,4}$, \\ 
\vspace{0.25em} Cheng Chi$^{2}$, Xiansheng Chen$^{2}$, Qiongyu Zhang$^{3}$, Zhongxia Zhao$^{1,2}$, \\
\vspace{0.25em} Pengwei Wang$^{2,\dagger}$, Zhongyuan Wang$^2$, Shanghang Zhang$^{1,2,\text{\Letter}}$ \\
$^1$ \small State Key Laboratory of Multimedia Information Processing, School of Computer Science, Peking University \\ 
\vspace{0.25em} $^2$ \small Beijing Academy of Artificial Intelligence,
$^3$ \small University of Sydney, 
$^4$ \small Institute of Automation, Chinese Academy of Sciences \\
\textit{\textbf{Project website:}} \href{https://action-sketcher.github.io/}{https://action-sketcher.github.io}
}

\begin{document}

\twocolumn[{%
\maketitle
\vspace{-0.9cm}
\begin{center}
    \centering
    \captionsetup{type=figure}
    \includegraphics[width=0.98\linewidth]{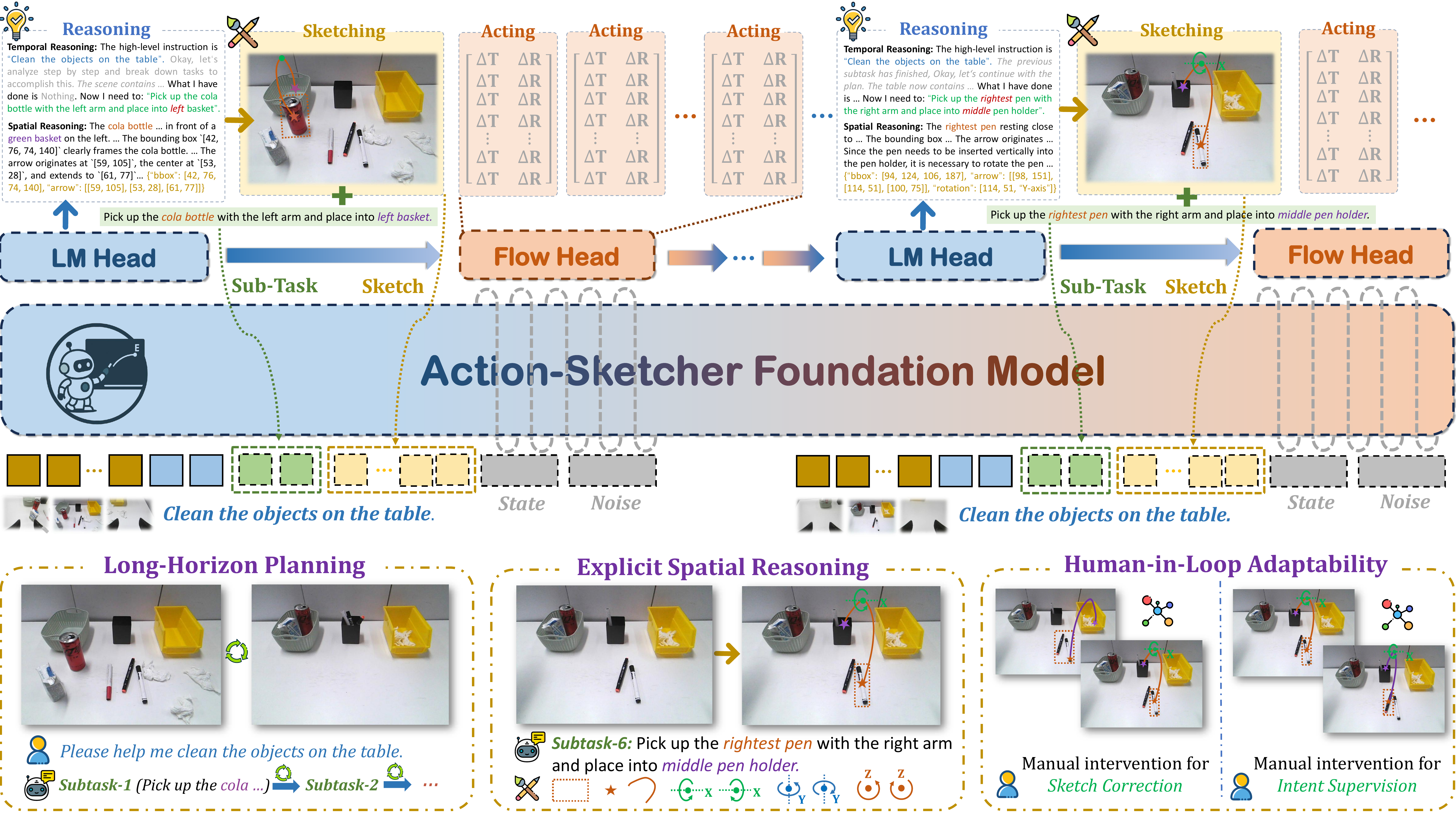}
    \captionof{figure}{
    \textbf{Overview of Action-Sketcher.} Our framework operates in a See-Think-Sketch-Act loop, where a foundation model first performs temporal and spatial reasoning to decompose a high-level instruction (e.g., ``Clean the objects on the table'') into a subtask and a corresponding \textit{Visual Sketch}. This sketch, composed of primitives like points, boxes, and arrows, serves as an explicit, human-readable plan that guides a low-level policy to generate robust action sequences. This methodology enables three key capabilities: \textit{\textbf{(bottom left)}} long-horizon planning through task decomposition, \textit{\textbf{(bottom middle)}} explicit spatial reasoning by grounding instructions in scene geometry, and \textit{\textbf{(bottom right)}} seamless human-in-the-loop adaptability via direct sketch correction and intent supervision.
    } 
    \label{fig:teaser}
\end{center}
}]

\let\thefootnote\relax\footnotetext{$^{*}$ Equal contribution.}
\let\thefootnote\relax\footnotetext{$^{\dagger}$ Project leaders.}
\let\thefootnote\relax\footnotetext{$^{\text{\Letter}}$ Corresponding author: \href{shanghang@pku.edu.cn}{shanghang@pku.edu.cn}}

\begin{abstract}
Long-horizon robotic manipulation is increasingly important for real-world deployment, requiring spatial disambiguation in complex layouts and temporal resilience under dynamic interaction. However, existing end-to-end and hierarchical Vision–Language–Action (VLA) policies often rely on text-only cues while keeping plan intent latent, which undermines \textit{referential grounding} in cluttered or underspecified scenes, impedes effective \textit{task decomposition} of long-horizon goals with close-loop interaction, and limits \textit{causal explanation} by obscuring the rationale behind action choices. To address these issues, we first introduce \textbf{Visual Sketch}, an implausible visual intermediate that renders points, boxes, arrows, and typed relations in the robot’s current views to externalize spatial intent, connect language to scene geometry, and provide a human-verifiable bridge between high-level reasoning and low-level control. Building on \textit{Visual Sketch}, we present \textbf{Action-Sketcher}, a VLA framework that operates in a cyclic \textit{See $\rightarrow$ Think $\rightarrow$ Sketch $\rightarrow$ Act} workflow coordinated by adaptive token-gated strategy for reasoning triggers, sketch revision, and action issuance, thereby supporting reactive corrections and human interaction while preserving real-time action prediction. To enable scalable training and evaluation, we curate diverse corpus with interleaved images, text, \textit{Visual Sketch} supervision, and action sequences, and train \textit{Action-Sketcher} with a multi-stage curriculum recipe that combines interleaved sequence alignment for modality unification, language-to-sketch consistency for precise linguistic grounding, and imitation learning augmented with sketch-to-action reinforcement for robustness. Extensive experiments on cluttered scenes and multi-object tasks, in simulation and on real-world tasks, show improved long-horizon success, stronger robustness to dynamic scene changes, and enhanced interpretability via editable sketches and step-wise plans. Project website: \href{https://action-sketcher.github.io/}{Action-Sketcher}.
\end{abstract}
    
\section{Introduction}
\label{sec:intro}

Robotic manipulation is moving beyond short, single-step primitives toward \textit{long-horizon, open-world} tasks in which goals, layouts, and human preferences evolve over time. In these settings, an agent must not only \textit{see} and \textit{act} but also maintain an interpretable decision chain that remains reliable under shifting spatial relations and temporal plans.

However, achieving such reliability is hindered by two primary bottlenecks. \textbf{\textit{(1) On the spatial side,}} language-to-action grounding is brittle: natural-language instructions are often \textit{ambiguous} (\textit{e.g.,} ``pour tea into the cup'' in a multi-cup scene) or \textit{underspecified} (\textit{e.g.,} ``place the book left of the mug,'' leaving the exact target pose undefined). To resolve referential uncertainty and translate linguistic relations into executable constraints, we advocate \textbf{\textit{explicit visual sketches}} (i.e., region highlights, keypoints, and relational arrows anchored to scene geometry). \textbf{\textit{(2) On the temporal side,}} human-in-the-loop coordination remains weak: real-time interaction is limited, and \textit{explainable planning} artifacts are rarely exposed, allowing small errors to propagate undetected. Making temporal structure explicit via step-wise rationales enables early error detection and efficient recovery. Together, these challenges highlight that deep reasoning paired with explicit visual sketches is essential for reliable manipulation in dynamic environments.

Recent Vision–Language–Action (VLA) models \cite{kim2024openvla, liu2024rdt, black2024pi_0, Helix2024, team2025gemini_robo, bjorck2025gr00t,bai2025towards} have made notable progress in mapping observations and language directly to actions. However, because plan intent is largely embedded in latent representations, these models struggle with task decomposition and causal explanation, both of which are critical for long-horizon operation in dynamic scenes. Hierarchical VLAs \cite{vemprala2024chatgpt, shi2025hi, pi05} attempt to address these challenges with planner and controller systems, yet their reasoning is often instantaneous and lacks persistent modeling of \textit{global intent} (\textit{e.g.} evolving human goals, emerging errors, and prior states); they also provide limited support for \textit{referential disambiguation}, leaving spatial references unresolved in cluttered environments. A complementary direction introduces \textit{think before act} reasoning within VLAs, for example EO-1 \cite{qu2025eo1} with interleaved vision, thinking, and action pretraining, and OneTwoVLA \cite{lin2025onetwovla} with adaptive joint reasoning and acting, but when the intermediate evidence remains \textit{text only}, the spatial referential disambiguation behind actions (contact points, approach directions, object relations) stays implicit, which hinders human verification and deprives the controller of low-entropy geometric guidance. 

To address these, we externalize spatial intent at the language to control interface by introducing an explicit \textit{visual intermediate representation} that we call \textbf{\textit{Visual Sketch}}, which is a sparse set of action oriented primitives rendered on the current view or views: \textit{points} (key contact or aim locations), \textit{bounding boxes} (object or region selection), and \textit{arrows} (desired motion or force directions). Co grounding language with these geometric sketches provides disambiguation of referents and spatial constraints, improves supervisability via simple annotation or auto labeling, and enables interpretability and debuggability because humans can read, approve, or modify intent before execution. In effect, the sketch serves as a verifiable contract between high level reasoning and low level control.

Built upon Visual Sketch, we propose \textbf{\textit{Action-Sketcher}}, a VLA framework that operates in a \textit{See $\rightarrow$ Think $\rightarrow$ Sketch $\rightarrow$ Act} loop, as also shown in Fig.~\ref{fig:teaser}. Given multi-view observations and a task instruction, the model alternates among (1) \textit{Think}, which produces a concise rationale with temporal planning and spatial conception; (2) \textit{Sketch}, which emits a scene- and subtask-tailored \textit{Visual Sketch} (i.e., points, boxes and arrows); and (3) \textit{Act}, which synthesizes a low-latency action chunk conditioned on the current observations. To enable adaptive switching and event-aware control, an adaptive token-gated strategy orchestrates the loop, \textit{e.g.,} 
\texttt{<BOR>} (begin-of-reasoning) initiates or refreshes global reasoning, and \texttt{<BOA>} (begin-of-action) triggers action-chunk prediction. 
The policy transitions among these modes based on observed state, predicted risk, and user feedback, thereby supporting reactive corrections and human interaction while preserving real-time action prediction. To support scalable training and evaluation, we curate a corpus that couples interleaved images, text, and action sequences with \textit{Visual Sketch} supervision across diverse manipulation skills, cluttered tabletops, and multi-object scenes. 
Our multi-stage curriculum training recipe combines (i) \textit{foundational spatio-temporal learning} for modality unification; (ii) \textit{language-to-sketch consistency} to ensure precise phrase-to-geometry bindings (\textit{e.g.} “grasp the handle” $\rightarrow$ handle keypoint + approach arrow in the \textit{Visual Sketch}); and (iii) imitation learning augmented with \textit{sketch-to-action reinforcement} for robustness.
The contributions of this paper are summarized as follows:
\begin{itemize}[left=1em]
    \item We formalize \textbf{\textit{Visual Sketch}} as a co-grounded, explicit spatial-intent interface that renders points, boxes and arrows, thereby disambiguating \emph{where} and \emph{how} to act and serving as a human-verifiable contract between high-level reasoning and low-level control.
    \item We propose \textbf{\textit{Action-Sketcher}}, a VLA framework that operates in a \textit{See $\rightarrow$ Think $\rightarrow$ Sketch $\rightarrow$ Act} loop, orchestrated by token-gated states for adaptive switching among reasoning, visual-sketch generation/revision, and action synthesis, enabling real-time interruption handling, error detection, and sketch-level correction without sacrificing latency.
    \item We curate an interleaved corpus and a training recipe that align language, \textit{Visual Sketches}, and actions via interleaved sequence alignment, language-to-sketch consistency, and imitation learning with sketch-to-action reinforcement; empirically, the resulting system improves long-horizon success, robustness to dynamic scene changes, and interpretability, validated through comprehensive ablations and real-robot studies.
\end{itemize}

\section{Related Work}
\label{sec:related_work}

\textbf{Vision--Language--Action Models (VLAs).}
VLAs connect multimodal perception to low-level control on top of multimodal LLM backbones. End-to-end policies employ regression-based~\cite{li2023vision, brohan2023rt, kim2024openvla, li2024lamp, hao2025tla, liu2024robomamba}, diffusion-based~\cite{liu2024rdt, li2025language, black2024pi_0, li2025you, li2025robomirror, team2024octo}, or hybrid mappings~\cite{liu2025hybridvla} from observations and language to actions, achieving strong short-horizon performance but often keeping plan intent latent, limiting \textit{task decomposition} and \textit{causal explanation}. Hierarchical VLAs~\cite{liu2024self, zhong2025dexgraspvla, Helix2024, team2025gemini_robo, bjorck2025gr00t, shi2025hi, pi05} introduce planner–controller splits to improve long-horizon behavior; however, reasoning is frequently instantaneous or local and lacks persistent modeling of global intent (\textit{e.g.} evolving human goals, emerging errors, and prior states), which hampers human-in-the-loop interaction and referential disambiguation in dynamic scenes. Recent \textit{think before act} variants integrate explicit reasoning within a unified backbone, such as \textsc{EO-1}~\cite{qu2025eo1} and \textsc{OneTwoVLA}~\cite{lin2025onetwovla}. \textsc{ThinkAct}~\cite{huang2025thinkact} further couples textual plans with a reinforced \emph{visual plan latent} to condition downstream action models, improving long-horizon planning; yet the plan is ultimately compressed into a latent that is not human-editable, limiting persistent, verifiable intent during execution. This motivates an interface that makes intent observable and actionable.

\textbf{Robotic visual prompting, sketches, and traces.}
Visual prompting encodes action intent via geometric or pictorial cues rendered on the image plane. At the representation level, \textsc{LLARVA}~\cite{niu2024llarva} predicts two-dimensional traces to align vision and action spaces, and \textsc{RoboBrain}~\cite{ji2025robobrain, team2025robobrain} emits explicit visual guidance (\textit{e.g.} pointing, affordances, trajectories) for downstream policies. At the policy level, \textsc{RT-Trajectory}~\cite{gu2023rt} conditions on rough trajectory sketches, \textsc{RT-Sketch}~\cite{sundaresan2024rt} leverages hand-drawn goal sketches under ambiguity and distractors, \textsc{TraceVLA}~\cite{zheng2024tracevla} injects visual trace prompting, and \textsc{RoVI}~\cite{li2025rovi} systematizes object-centric symbols (arrows, circles, colors, numbers). Contemporaneously, \textsc{MolmoAct}~\cite{lee2025molmoact} generates depth-aware tokens and mid-level spatial plans as editable trajectory traces to steer actions. While these works show the value of explicit visual cues, they either freeze prompts as static inputs or compress plans into non-editable latents. In contrast, our Visual Sketch is persistent, human-editable, and produced one sub-task at a time, providing targeted and scene-aligned guidance. Because sketches no longer encode the entire trajectory, they can express richer and more precise action primitives such as contact keypoints, rotation arrows, and placement cues instead of coarse traces.

\section{Method}
\label{sec:method}

Our proposed framework, Action-Sketcher, is designed to address the challenges of spatial ambiguity and temporal brittleness in long-horizon manipulation tasks. By introducing an explicit visual intermediate, we create a verifiable and correctable interface between high-level reasoning and low-level action execution. This section first details the structure and semantics of our proposed \textit{Visual Sketch} in Sec.\ref{sec:visual-sketch}. We then present the complete \emph{Action–Sketcher} framework in Sec.\ref{sec:arch}, encompassing the problem formulation, the dual-mode inference pipeline, and the multi-stage curriculum learning strategy.

\begin{figure*}[t]
    \centering
    \vspace{-0.2cm}
    \includegraphics[width=0.98\linewidth]{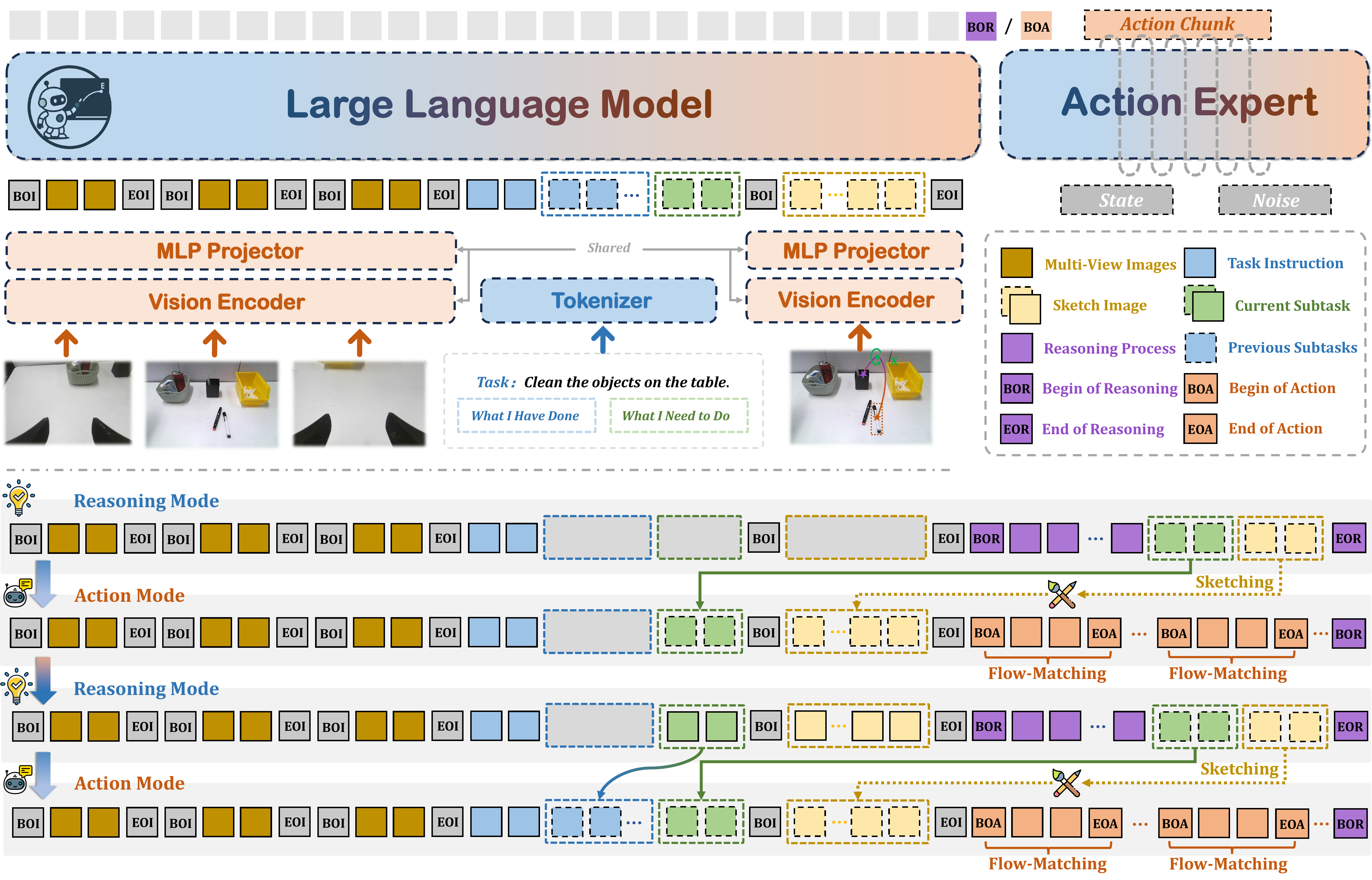}
    \caption{\textbf{Overview of Action\textendash Sketcher.}
    The model runs an event–driven loop that (i) summarizes the next subtask, (ii) emits a compact \emph{Visual Sketch} (points, boxes, arrows, relations) that externalizes spatial intent, and (iii) synthesizes an action chunk conditioned on that sketch and the robot state. The explicit intermediate supports targeted supervision, on–the–fly correction, and reliable long–horizon execution within a single–model architecture.}
    \label{fig:method}
\end{figure*}

\subsection{Visual Sketch: Explicit Spatial Intent}
\label{sec:visual-sketch}
During task execution, the agent receives multi-view observations and a task instruction. To promote broad adaptability across platforms and scenes, we adopt the robot's ego-view camera as the canonical reference frame in which the \emph{Visual Sketch} is expressed; all primitives are defined on this ego-view image plane. At each step $t$, the \emph{Visual Sketch} is a sparse tuple of geometric primitives that explicitly represents the spatial intent:
\begin{equation}
\mathcal{S}_t=\big(\mathcal{B}_t,\ \mathcal{P}_t,\ \mathcal{A}_t\big),
\end{equation}
\noindent where $\mathcal{B}_t$ designates target regions with object-level bounding boxes, $\mathcal{P}_t$ collects keypoints critical for task completion, and $\mathcal{A}_t$ specifies intended motions with arrows defined over the keypoints in $\mathcal{P}_t$. Each primitive is defined as follows:

\textit{\textbf{Boxes} ($\mathcal{B}_t$)}.
Bounding boxes serve as object-level affordance cues, grounding the task by delineating regions of target object which the robot is permitted to operate or interact. This disambiguates object references in cluttered scenes and abstracts away fine-grained appearance while preserving essential scale and placement information:
\begin{equation}
\mathcal{B}_t=\{b_i\}_{i=1}^{N_b}, \qquad b_i=(x_{1,i}, y_{1,i}, x_{2,i}, y_{2,i}),
\end{equation}
which is parameterized by the pixel coordinates of its top-left $(x_{1,i}, y_{1,i})$ and bottom-right $(x_{2,i}, y_{2,i})$ corners. For instance, given the instruction \textit{``pick up the item closest to the cup,''} a bounding box can be generated to unambiguously designate the target object (\textit{e.g.} an apple) by resolving the spatial relation.

\textit{\textbf{Points} ($\mathcal{P}_t$)}.
We use keypoints to specify precise locations where the robot should interact with or move relative to objects. These points can represent part-level affordances, motion waypoints, or geometric reference points. Formally, this set is expressed as:
\begin{equation}
\mathcal{P}_t=\{p_i\}_{i=1}^{N_p}, \qquad p_i=(x_i,y_i),
\end{equation}
where $(x_i,y_i)$ are pixel coordinates on the ego-view image plane.
Taking \emph{``tea pouring''} as an example, we might denote the teapot spout tip by \(p_{\mathrm{spout}}\), the cup's center by \(p_{\mathrm{cup}}\), and a stable contact point on the handle by \(p_{\mathrm{handle}}\). First, to \emph{``lift the teapot''}, the action policy can focus on \(p_{\mathrm{handle}}\) as a robust grasp anchor, representing a part-level affordance. After lifting, to \emph{``move the teapot towards the cup''}, the pair \((p_{\mathrm{spout}}, p_{\mathrm{cup}})\) serves as a geometric reference for alignment, where $p_{\mathrm{spout}}$ is the start waypoint and $p_{\mathrm{cup}}$ is the target. 
These keypoint anchors enable the model to comprehend and execute precise geometric subtasks.

\textbf{Arrows} ($\mathcal{A}_t$).
Arrows serve as the dynamic link between static keypoints and actuation. We posit that complex $SE(3)$ manipulation can be effectively factorized into translational trajectories and rotational cues projected onto the 2D image plane. Anchored to the keypoints $\mathcal{P}_t$, the arrow set $\mathcal{A}_t$ encodes these intended motions as the union of two subsets: $\mathcal{A}_t = \mathcal{A}^{\mathrm{trans}}_t \cup \mathcal{A}^{\mathrm{rot}}_t$.
\emph{(i) Translation Arrows} encode the intended trajectory of the end-effector:
\begin{equation}
\mathcal{A}^{\mathrm{trans}}_t=\{a^{\mathrm{trans}}_i\}_{i=1}^{N_{trans}}, \quad a^{\mathrm{trans}}_i=\big(p_i^{\mathrm{start}}, p_i^{\mathrm{end}}\big),
\end{equation}
where each arrow is an ordered sequence anchored to keypoints in $\mathcal{P}_t$. In the \emph{``tea pouring''} example, a primary arrow ($p_{\text{spout}}\!\rightarrow\!p_{\text{cup}}$) guides the teapot spout towards the cup center. Subsequently, short corrective arrows may be generated to recenter the spout if drift is observed. After pouring, a retraction arrow moves the spout away to prevent dripping.
\emph{(ii) Rotation Arrows} specify rotations about canonical axes:
\begin{equation}
\mathcal{A}^{\mathrm{rot}}_t=\{a^{\mathrm{rot}}_i\}_{i=1}^{N_{rot}},
\end{equation}
\begin{equation}
a^{\mathrm{rot}}_i=\big(p_i,\ \text{axis}\in\{x,y,z\},\ \text{dir}\in\{\circlearrowright,\circlearrowleft\}\big),
\end{equation}
where each arrow defines a rotation centered at $p_i\in \mathcal{P}_t$ about a canonical axis ($x, y, z$) in a specified direction (clockwise $\circlearrowright$ or counter-clockwise $\circlearrowleft$). In our \emph{``tea pouring''} example, a rotation arrow centered near $p_{\text{handle}}$ around the ego-view x-axis (horizontal) would instruct the tilt required for pouring. A subsequent counter-rotation would level the teapot. Similarly, a rotation around the z-axis (perpendicular to the image plane) could be used to adjust the spout's yaw orientation.

\subsection{Action Sketcher Framework}
\label{sec:arch}

Our framework formulates long-horizon manipulation as a sequence modeling task over a hybrid output space (\textit{i.e.,} discrete and continuous). We aim to learn a policy $\pi_\theta$ that maps context to a joint output distribution, enabling the agent to seamlessly interleave discrete reasoning (for planning and sketching) and continuous action prediction, autonomously deciding the appropriate mode at each timestep.

\subsubsection{See-Think-Sketch-Act Pipeline}

As illustrated in Fig.~\ref{fig:method}, the model's input context is a sequence of tokens representing multi-view images (\textit{e.g.} left-wrist, right-wrist, base), the task instruction, a history of completed subtasks, the current subtask, and the visual sketch image. Action-Sketcher operates in two distinct, adaptively switched modes: \textit{Reasoning Mode} and \textit{Action Mode}. This dual-mode design allows the model to dynamically balance the need for careful deliberation with the demand for real-time execution.
The switch between modes is governed by a token-gated mechanism. 

\textbf{\textit{(1) Reasoning Mode.}} If the model determines that reasoning is required (\textit{e.g.} after completing a subtask, encountering an error, or receiving human intervention), it generates a special token, \texttt{<BOR>} (begin-of-reasoning). Following this, the model auto-regressively performs temporal and spatial reasoning.
\textit{Temporal Reasoning:} The model first analyzes the current scene, considering the overall task instruction and the history of completed subtasks to deduce the next logical subtask.
\textit{Spatial Reasoning:} Based on the newly identified subtask, the model performs spatial reasoning about the object layout and relationships in the scene. This process yields the text-based elements for the \textit{Visual Sketch} (points, boxes, arrows) that correspond to the subtask.
This reasoning phase concludes with the generation of an \texttt{<EOR>} (end-of-reasoning) token. The generated text-based sketch is then rendered onto the current reference view to create an image-based \textit{Visual Sketch}. The input context is subsequently updated with the new subtask and the new sketch image.

\textbf{\textit{(2) Action Mode.}} Conversely, if the model deems reasoning unnecessary, such as during the routine execution of a subtask where the scene remains consistent, it generates a \texttt{<BOA>} (begin-of-action) token. This triggers the action expert to generate action chunks via flow-matching~\cite{lipman2022flow}.

Initially, the context tokens for completed subtasks, the current subtask, and the visual sketch image are empty. The model must begin in \textit{Reasoning Mode} to populate these fields. Subsequently, it can fluidly switch between modes to maintain both accuracy and real-time performance.

\subsubsection{Model Structure}

The Action-Sketcher framework is model-agnostic and can be integrated with any VLA model. For a specific instance, we employ $\pi_0$~\cite{black2024pi_0} as the backbone. The model is trained to auto-regressively generate textual reasoning chains, subtask plans, and structured descriptions of visual sketches, while simultaneously optimizing a flow-matching loss~\cite{lipman2022flow} to predict continuous action chunks. Details for model structure can refer to Appendix~\ref{app:modelstructure}.

\subsubsection{Training Strategy}

To realize the ``See-Think-Sketch-Act'' pipeline, we employ a multi-stage curriculum learning strategy. This curriculum is designed to first establish a broad foundation of spatiotemporal understanding (Stage 1), then specialize this knowledge to align high-level language with our structured \textit{Visual Sketch} representation (Stage 2), and finally, to jointly train the action policy and the adaptive mode-switching mechanism for robust, real-world execution (Stage 3). We elaborate on each stage below.

\textbf{Stage 1: Foundational Spatiotemporal Learning.}
The initial stage focuses on pre-training the model to develop general spatiotemporal modeling and reasoning capabilities. We utilize a large-scale dataset for this purpose. For spatial understanding, the training data comprises 3.4M samples from diverse sources, including Visual Grounding datasets~\cite{gupta2019lvis, krishna2017visual, ji2025robobrain, li2024llavaov} for object-level bounding box prediction, Spatial Pointing datasets~\cite{zhou2025roborefer, team2025robobrain, yuan2024robopoint, deitke2024molmo, wang2025towards} for keypoint prediction, Scene Understanding datasets~\cite{mmscan,3rscan,scanqa} for broader scene perception, and general visual question-answering (VQA) datasets~\cite{liu2023mitigating, liu2024improved,liu2024improved}. For temporal learning, we construct a dataset of 870k sequences by integrating data from EgoPlan~\cite{chen2023egoplan}, ShareRobot~\cite{ji2025robobrain}, and AgiBot-World~\cite{bu2025agibot}. To foster explicit reasoning, we annotate 20\% of this corpus with detailed textual rationales generated by GPT-4o~\cite{hurst2024gpt4o}. This stage equips the Action-Sketcher with robust spatiotemporal reasoning and instruction-following abilities.

\textbf{Stage 2: Reasoning-to-Sketch Enhancement.}
In this stage, the model learns the coherent reasoning process required for the Reasoning Mode. The objective is to analyze a scene and, guided by the task instruction and execution history, perform sequential temporal and spatial reasoning to generate the next subtask and its corresponding text-based \textit{Visual Sketch}. The training data is curated from 2.6k episodes of diverse, long-horizon (with 2--16 subtasks) manipulation tasks in complex layouts (over 20 object types with various spatial arrangements) that we collected in the real world, encompassing tasks such as tidying a tabletop, pouring tea, and general pick-and-place scenarios. We further supplement this by annotating 1.7k full trajectories from existing datasets, including LIBERO~\cite{liu2023libero} and RoboTwin~2.0~\cite{chen2025robotwin}. This culminates in a reasoning-to-sketch dataset of 21k samples used to fine-tune the Action-Sketcher to master the complete Reasoning Mode pipeline.

\textbf{Stage 3: Sketch-to-Action and Mode Adaptation.}
The final stage jointly trains the action policy and the mode-switching mechanism. The training data serves two purposes. First, it teaches adaptive mode selection by providing examples where a \texttt{<BOR>} token is appropriate (\textit{e.g.} at subtask boundaries or upon scene changes) versus a \texttt{<BOA>} token (during nominal execution). Second, it trains the action expert using action-labeled data. To enhance robustness, we augment the \textit{Visual Sketch} corresponding to each action label to simulate potential inference-time inaccuracies. Specifically, for \textit{bounding boxes}, we apply random perturbations to the ground-truth box while maintaining an Intersection over Union (IoU) of at least 0.8. For \textit{points}, we randomly sample new coordinates within a circle of a small radius $c$ around the ground-truth point. \textit{Arrows} are adjusted according to \textit{points}. This augmentation ensures the action expert is resilient to minor errors in the generated sketch.

Furthermore, we address the significant data imbalance inherent in task execution, where Action Mode steps far outnumber Reasoning Mode steps. To prevent the model from developing a bias towards the more frequent \texttt{<BOA>} token, we employ a mode-balanced sampling strategy. Let $D_R$ and $D_A$ be the sets of training samples for Reasoning Mode and Action Mode. The probability $P(d)$ of selecting a sample $d$ from the combined dataset $D = D_R \cup D_A$ is given by:
\begin{equation}
P(d) = \begin{cases}
    \frac{1}{2|D_R|} & \text{if } d \in D_R \\
    \frac{1}{2|D_A|} & \text{if } d \in D_A
\end{cases}
\label{eq:sampling}
\end{equation}
This approach directly mitigates the data imbalance, preventing the model from developing a mode bias and fostering a more robust mode-selection capability.

\section{Experiments}
\label{sec:exp}

\begin{figure*}[t]
\centering
\includegraphics[width=1.0\linewidth]{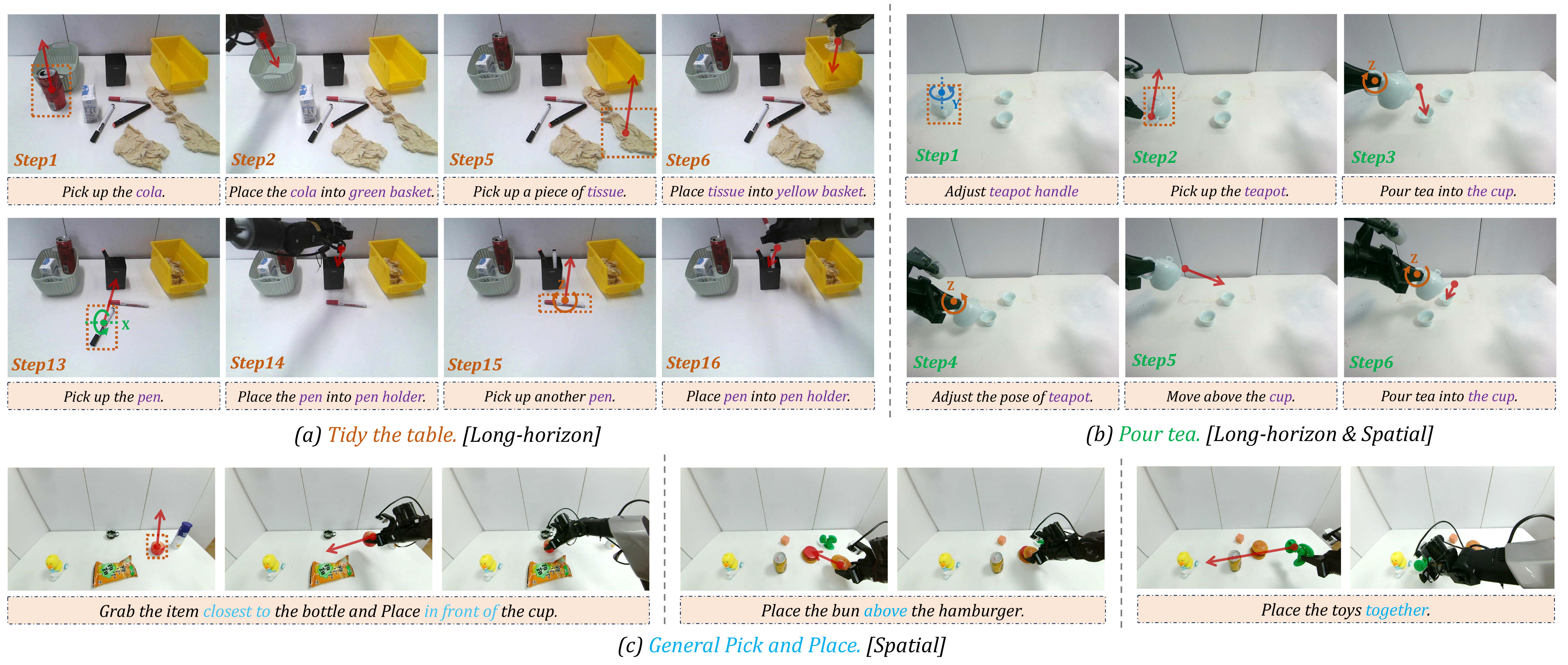}
\caption{\textbf{Real-World Demonstrations of Action-Sketcher.} Qualitative rollouts on long-horizon and spatial manipulation tasks. Our framework generates explicit \textit{Visual Sketches} (overlaid points, boxes, and arrows) to ground high-level reasoning into low-level actions, successfully completing tasks like tidying a tabletop and pouring tea in cluttered environments.}
\label{fig:qual}
\end{figure*}

We conduct extensive experiments on both simulated environments and a real-world robotic platform to validate the effectiveness of Action-Sketcher in long-horizon, complex manipulation tasks. Our evaluation is structured around three key research questions:
\begin{itemize}[leftmargin=*]
    \item \textbf{RQ1:} How does the performance of Action-Sketcher, across tasks of varying temporal and spatial complexity, compare to state-of-the-art baseline models?
    \item \textbf{RQ2:} What are the primary failure modes of our system, and how effectively can human-in-the-loop intervention, via the \textit{Visual Sketch} interface, improve the success rate?
    \item \textbf{RQ3:} How do the core components of the Action-Sketcher framework affect the performance?
\end{itemize}

\subsection{Task Performance Evaluation (RQ1)}
\label{sec:exp_rq1}

\textbf{Experiment Settings.} We first evaluate the overall task success rate of Action-Sketcher against a suite of VLA baselines.
Our evaluation is performed in two simulated benchmarks and on a real-world robotic arm. For simulation, we use LIBERO~\cite{liu2023libero}, a benchmark focused on lifelong skills, and an enhanced version of RoboTwin~2.0~\cite{chen2025robotwin} with increased object clutter and spatial complexity, including \textit{``stack three blocks''}, \textit{``hanging mugs''}, \textit{``place empty cups''} and \textit{``place a2b anywhere''}. For the real world, we evaluate on three long-horizon tasks: \textit{``tidying a cluttered tabletop''}, \textit{``pouring tea''}, and \textit{``general pick-and-place''} with ambiguous instructions, built upon both Agilex and Galaxea dual-arm robot platform. We report the average success rate across all tasks in each benchmark suite.

\textbf{Baseline.} We compare our model against several categories of the VLA models: End-to-End VLAs (DP~\cite{chi2023diffusion}, Octo~\cite{team2024octo}, OpenVLA~\cite{kim2024openvla}), VLAs with specialized architectures (SpatialVLA~\cite{qu2025spatialvla}, $\pi_0$~\cite{pi05}, $\pi_{0.5}$~\cite{pi05}, OpenVLA-OFT~\cite{kim2025fine}), and models that incorporate forms of visual prompting or intermediate representations (TraceVLA~\cite{zheng2024tracevla}, Molmo-ACT~\cite{lee2025molmoact}, PixelVLA~\cite{liang2025pixelvla}).

\begin{table}[t]
\centering
\caption{Success rates (\%) on the LIBERO benchmark. The best results among different models are highlighted in \textbf{bold}, while the second-best results are \underline{underlined}.}
\label{tab:main_results}
\resizebox{\columnwidth}{!}{%
\begin{tabular}{lccccc}
\toprule
\textbf{Method} & \textbf{Spatial} & \textbf{Object} & \textbf{Goal} & \textbf{Long} & \textbf{Avg.} \\
\midrule
Diffusion Policy~\cite{chi2023diffusion} & 78.3 & 92.5 & 68.3 & 50.5 & 72.4 \\
Octo~\cite{team2024octo} & 78.9 & 85.7 & 84.6 & 51.1 & 75.1 \\
OpenVLA~\cite{kim2024openvla} & 84.7 & 88.4 & 79.2 & 53.7 & 76.5 \\
SpatialVLA~\cite{qu2025spatialvla}	& 88.2 & 89.9 & 78.6 & 55.5 & 78.1 \\
$\pi_0$ + FAST~\cite{pertsch2025fast} & 96.4 & 96.8 & 88.6 & 60.2 & 85.5 \\
$\pi_0$~\cite{pi05} & 96.8 & \underline{98.8} & 95.8 & 85.2 & 94.1 \\
$\pi_{0.5}$~\cite{pi05} &  \textbf{98.8} & 98.2 & \textbf{98.0} & 92.4 & 96.8 \\
OpenVLA-OFT~\cite{kim2025fine} &  \textbf{97.6} & 98.4 & \underline{97.9} & \underline{94.5} & \textbf{97.1} \\
\midrule
\multicolumn{6}{l}{\textbf{\textit{VLAs with Visual Prompts}}} \\
\midrule
TraceVLA~\cite{zheng2024tracevla} & 84.6 & 85.2 & 75.1 & 54.1 & 74.8 \\
Molmo-ACT~\cite{lee2025molmoact} & 87.0 & 95.4 & 87.6 & 77.2 & 86.6 \\
PixelVLA~\cite{liang2025pixelvla} & 88.5 & 90.0 & 85.8 & 82.6 & 86.7 \\
\rowcolor[HTML]{DAEFF9} \textbf{Action-Sketcher (Ours)} &97.2 & \textbf{99.6} & 94.8 & \textbf{96.0} & \underline{96.9} \\
\bottomrule
\end{tabular}%
}
\vspace{-1em}
\end{table}

\begin{table*}[h]
\centering
\caption{Success rates (\%) on selected long-horizon and spatially complex tasks, comparing Action-Sketcher to top baselines. For simulation scores, the success rate is based on task completion, while for real-world execution, the success rate is based on the average subtask completion rate on all rollouts. The best results among different models are highlighted in \textbf{bold}, while the second-best results are \underline{underlined}.}
\label{tab:long_horizon_tasks}
\resizebox{1.0\textwidth}{!}{%
\begin{tabular}{l|ccccc|ccc}
\toprule
& \multicolumn{5}{c|}{\textbf{\textit{RoboTwin 2.0 (Simulation)}}} & \multicolumn{3}{c}{\textbf{\textit{Real-World Execution}}} \\
 \cmidrule(lr){2-6} \cmidrule(lr){7-9}
\textbf{Method} & Stack Blocks & Hang Mugs & Empty Cups & Place A2B Left & Place A2B Right & Tidy Table & Pour Tea & Pick \& Place \\
\midrule
$\pi_0$~\cite{black2024pi_0} & 4.0 & 20.0 & 11.5 & 12.0 & 8.0 & 23.0 & 16.0 & 30.0 \\
$\pi_{0.5}$~\cite{pi05} & 7.0 & \textbf{27.0} & \underline{16.0} & 11.0 & 13.0 & 31.2 & \underline{20.0} & 34.5 \\
OpenVLA-OFT~\cite{kim2025fine} & \underline{12.4} & 23.5 & 13.5 & \underline{21.0} & \underline{15.0} & \underline{36.0} & 15.0 & \underline{52.5} \\
\midrule
\rowcolor[HTML]{DAEFF9} \textbf{Action-Sketcher (Ours)} & \textbf{34.5} & \underline{25.0} & \textbf{28.0} & \textbf{43.0} & \textbf{28.0} & \textbf{52.0} & \textbf{27.6} & \textbf{67.0} \\
\bottomrule
\end{tabular}%
}
\end{table*}

\textbf{Result.} As shown in Tab.~\ref{tab:main_results}, Action-Sketcher demonstrates superior performance across all evaluation suites. Notably, its advantage is most pronounced in the ``Long'' categories of the LIBERO benchmark, which specifically test for long-horizon planning. This suggests that the explicit See-Think-Sketch-Act cycle provides a significant advantage over methods that rely on latent planning.
To further analyze on long-horizon and spatially complex tasks, we specifically probe the capabilities of our model in scenarios requiring extended planning and intricate spatial reasoning, we present a focused comparison on four challenging tasks from the RoboTwin~2.0~\cite{chen2025robotwin} environment and our three real-world setups. These seven tasks were selected as they exemplify the primary challenges our work aims to address: long-term temporal dependencies and complex, often ambiguous, spatial relationships between objects. The results presented in Tab.~\ref{tab:long_horizon_tasks}, are unequivocal. While strong baselines like $\pi_{0.5}$~\cite{pi05} and OpenVLA-OFT~\cite{kim2025fine} perform competently, Action-Sketcher establishes a significant and consistent performance advantage across all tasks. This substantial performance margin validates our central hypothesis: for tasks where the plan is long and the spatial goals are complex, implicitly embedding intent within a latent space is insufficient. The explicit See-Think-Sketch-Act pipeline in Action-Sketcher empowers the agent to robustly decompose problems, ground its actions in an interpretable spatial representation, and execute with higher precision, making it effective in these challenging domains.

\subsection{Error Analysis and Human-in-Loop (RQ2)}
\label{sec:exp_rq2}

\begin{figure}[t]
\centering
\includegraphics[width=1.0\linewidth]{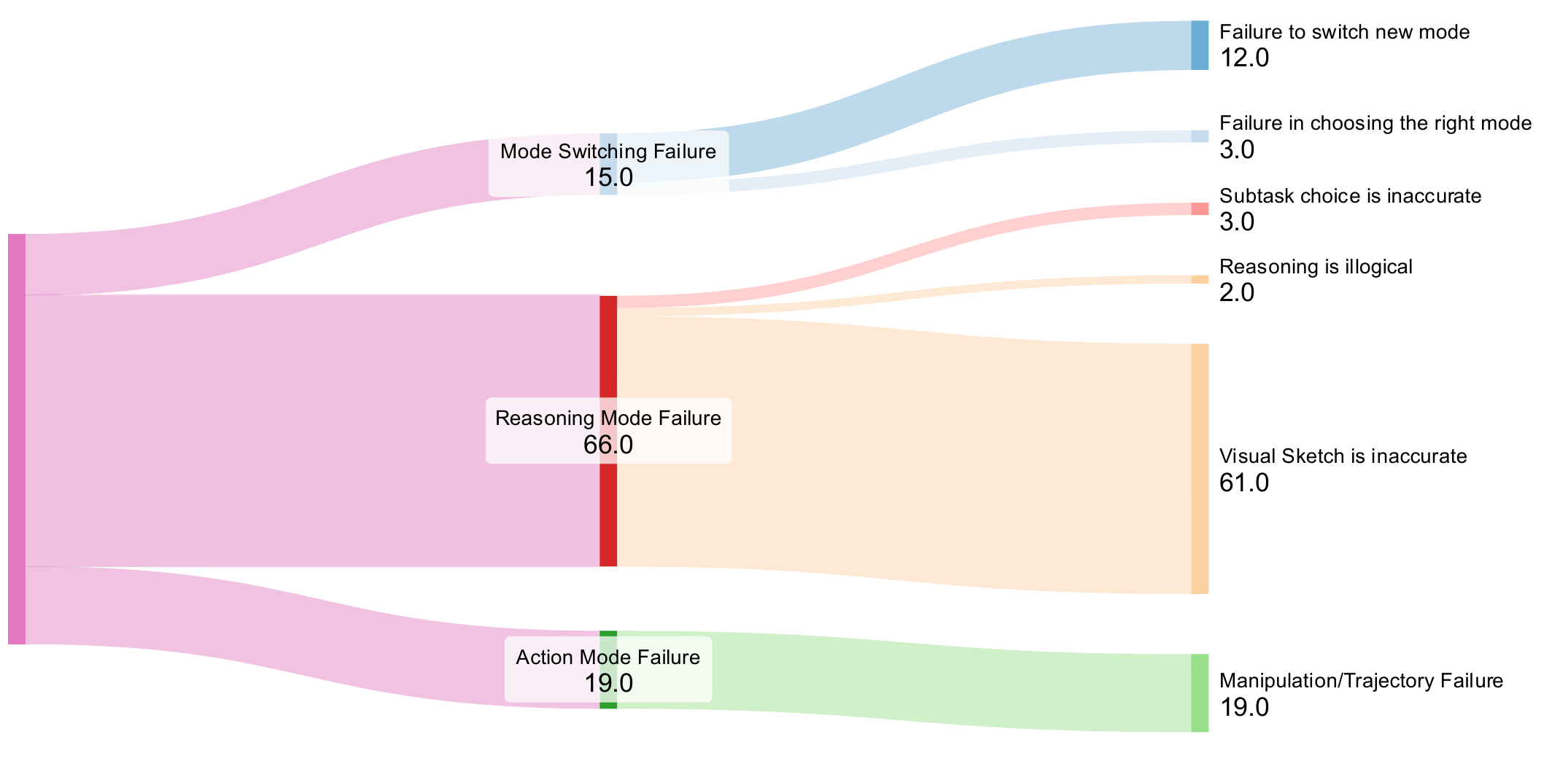}
\caption{Failure analysis of Action-Sketcher. Most errors arise in the Reasoning Mode, primarily due to inaccurate Visual Sketch generation.}
\label{fig:error}
\end{figure}

To better understand the limitations of our system and the potential for human collaboration, we analyze the root causes of task failures, as shown in Fig.~\ref{fig:error}. Our analysis reveals that failures can be attributed to three main sources: incorrect mode switching (12\% of failures), errors in the Reasoning Mode (66\%), and failures in the Action Mode (19\%). Within the Reasoning Mode, the vast majority of errors stem from the spatial reasoning phase—specifically, the generation of the \textit{Visual Sketch} (61\% of all failures).

This analysis highlights that inaccurate spatial grounding via the \textit{Visual Sketch} is the primary bottleneck. However, because the sketch is an explicit and human-interpretable interface, it presents a natural point for human intervention. To test this, we conducted an experiment where a human supervisor could pause execution and make minor edits to the generated sketch before the action was executed. As shown in Tab.~\ref{tab:human_in_loop}, this sketch-level correction mechanism dramatically improves performance, pushing the success rate on our most challenging real-world tasks to near-perfection. This demonstrates that Action-Sketcher not only performs well autonomously but also serves as an effective framework for human-robot collaboration.

\begin{table}[t]
\centering
\caption{Subtask completion rates (\%)  on various long-horizon real-world tasks with and without human-in-the-loop correction of the \textit{Visual Sketch}.}
\label{tab:human_in_loop}
\resizebox{0.48\textwidth}{!}{%
\begin{tabular}{lcc|c}
\toprule
\textbf{Real-World Task} & \textbf{Original $\uparrow$ (\%)} & \textbf{+ Human-in-Loop $\uparrow$ (\%)} & \textbf{$\Delta$ $\uparrow$ (\%)}\\
\midrule
Tidy Table & 52.0 & 75.0 & +23.0 \\
Pour Tea & 27.6 & 44.0 & +16.4\\
Pick \& Place & 67.0 & 85.5 & +18.5 \\
\bottomrule
\end{tabular}}
\end{table}

\subsection{Ablation Studies (RQ3)}
\label{sec:exp_rq3}
To assess the contribution of each system component, we conduct ablation studies across three dimensions, as detailed in Tab.~\ref{tab:ablation}.
\textit{\textbf{(1) Framework Components.}} Removing \textit{Spatial Reasoning} significantly degrades success to 13.8\%, while eliminating the \textit{Visual Sketch} entirely drops it further to 9.8\% (simulation) and 15.0\% (real-world). This confirms that the sketch is not merely an auxiliary visualization but a fundamental bridge for grounding language into executable actions.
\textit{\textbf{(2) Visual Primitives.}} While all primitives operate synergistically, \textit{Keypoints} prove most critical (26.6\%) by providing precise coordinate grounding. \textit{Arrows} and \textit{Bounding Boxes} are also essential for dynamic guidance and disambiguation; omitting either noticeably reduces reliability compared to the full model (34.5\%).
\textit{\textbf{(3) Training Curriculum.}} The results validate our multi-stage strategy. Skipping foundational pre-training (Stage 1) causes a moderate drop (29.2\%), whereas omitting reasoning fine-tuning (Stage 2) leads to a sharp decline (18.1\%) due to incoherent sketches. Crucially, removing the final adaptation (Stage 3) results in total failure (0.0\%), proving that Stage 3 is indispensable for the policy to interpret sketches to actions.

\begin{table}[t]
\centering
\caption{Ablation study on framework components, visual primitives, and training strategies. We report results on representative tasks: \textit{Stack Blocks} (Sim.) and \textit{Tidy Table} (Real). Success Rate indicates task completion in simulation, while Completeness measures the percentage of subtasks achieved in the real world.}

\label{tab:ablation}
\resizebox{\columnwidth}{!}{%
\begin{tabular}{lcc}
\toprule
\textbf{Model Variant} & \textbf{Success Rate $\uparrow$ (\%)} & \textbf{Completeness $\uparrow$ (\%)} \\
\midrule
\textit{\textbf{Action-Sketcher}} & \textbf{34.5} & \textbf{52.0} \\
\midrule
\rowcolor[HTML]{F2F2F2} \multicolumn{3}{l}{\textit{Framework Component Ablations}} \\
\midrule
w/o Spatial Reasoning & 13.8 & 23.9  \\
w/o Visual Sketch & 9.8 & 15.0 \\
\midrule
\rowcolor[HTML]{F2F2F2} \multicolumn{3}{l}{\textit{Visual Sketch Element Ablations}} \\
\midrule
w/o Bounding Box & 31.2 & 49.0 \\
w/o Keypoint & 26.6 & 43.6 \\
w/o Arrow & 29.9 & 48.2 \\
\midrule
\rowcolor[HTML]{F2F2F2} \multicolumn{3}{l}{\textit{Training Stage Ablations}} \\
\midrule
w/o Stage 1 & 29.2 & 39.7 \\
w/o Stage 2 & 18.1 & 21.9 \\
w/o Stage 3 & 0.0 & 0.0 \\
\bottomrule
\end{tabular}%
}
\vspace{0.5em}
\end{table}

\section{Conclusion}
\label{sec:conclusion}

In this work, we presented Action-Sketcher, a VLA framework designed to address spatial ambiguity and temporal brittleness in long-horizon robotic manipulation. To achieve this, we introduced \textit{Visual Sketch}, an explicit spatial representation that operationalizes the reasoning process through a structured See-Think-Sketch-Act cycle. By adaptively switching between reasoning to generate interpretable visual plans and executing actions conditioned on them, our approach outperforms state-of-the-art methods, particularly in tasks requiring complex spatial grounding and multi-step sequential planning. Extensive experiments and ablations further demonstrate that \textit{Visual Sketch} is the pivotal component, providing a robust bridge between high-level reasoning and low-level control. Its explicit nature not only improves autonomous performance but also enables effective human-in-the-loop correction, paving the way for more transparent and collaborative robotic agents. Future potential directions are detailed in Appendix.~\ref{future}.
\clearpage
{
    \small
    \bibliographystyle{ieeenat_fullname}
    \bibliography{main}
}
\clearpage
\maketitlesupplementary
\newpage
\appendix
\section*{Appendix}
This supplementary material provides more details of the proposed method and experiment results that are omitted from the manuscript due to the page limit. 
Sec.~\ref{sec1} presents additional details of the models, training strategies and evaluation settings. 
Sec.~\ref{sec2} presents details of training dataset.
Sec.~\ref{sec2_ext} presents the pipeline of data construction.
Sec.~\ref{sec3} complements more experiment results and analysis. 
Sec.~\ref{sec4} shows more visualization results to
prove the effectiveness of Action-Sketcher.
Sec.~\ref{future} discusses potential future research directions and limitations for Action-Sketcher.

\section{Details of Model, Training and Evaluation}
\label{sec1}
In this section, we provide comprehensive implementation details of the Action-Sketcher framework. We first elaborate on the specific model architecture based on the $\pi_0$ VLA policy and define the distinct input-output flows that facilitate the cyclic See-Think-Sketch-Act loop. Subsequently, we specify the detailed hyperparameters, computational configurations, and optimization strategies utilized across our three-stage curriculum training.

\subsection{Model Structure}
\label{app:modelstructure}
The Action-Sketcher framework is model-agnostic and can be integrated with any Vision-Language-Action (VLA) model. For our implementation and experiments, we employ $\pi_0$~\cite{black2024pi_0} as the base VLA policy. $\pi_0$ is a generalist robot policy that leverages a VLM backbone for semantic understanding and a Flow Matching policy for continuous action generation, which consists of three primary components:

 \begin{itemize}
    \item \textit{\textbf{Vision Encoder:}} a pre-trained Vision Transformer (\textit{i.e.,} SigLIP-SO400M~\cite{zhai2023sigmoid}) to process visual inputs. This encoder maps high-resolution images from multiple camera views (\textit{e.g.,} wrist and ego-centric views) and the rendered \textit{Visual Sketch} image into a sequence of visual tokens.
    
    \item \textit{\textbf{Auto-Regressive LLM Backbone:}} The core reasoning unit is a Transformer-based LLM (\textit{i.e.,} Gemma-3B). It takes a concatenated sequence of tokens as input, including the visual tokens, tokenized task instructions, and text-based history sub-tasks tokens. This backbone operates auto-regressively to perform high-level reasoning, generate text-based sketch parameters and sub-task description in the Reasoning Mode.
    
    \item \textit{\textbf{Flow Matching Action Expert:}} For low-level control (\textit{i.e.,} Action Mode), the model utilizes a Flow Matching head. Instead of auto-regressively discretizing actions, this head acts as a conditional generative model. It takes the robot's proprioceptive state and the latent embeddings from the LLM backbone as conditioning to generate continuous, high-frequency action chunks via solving a differential equation (ODE).
 \end{itemize}

\subsection{Inputs and Outputs}
Based on the VLA model, Action-Sketcher operates in a cyclic loop with distinct input-output flows depending on the active mode (\textit{i.e.,}  Reasoning Mode and Action Mode). The inputs and outputs are as follows:

\begin{itemize}
    \item \textit{\textbf{Reasoning Mode Inputs:}}
    The model receives:
    (1) Multi-view RGB images $I^{1:n}_t$ capturing the current scene (\textit{i.e.,} left-wrist, left-wrist and ego-centric views);
    (2) The task instruction $\ell$ (\textit{e.g.,}``Clean the objects on the table'');
    (3) Textual history of completed sub-tasks.
    (4) If the current sub-task is not the first in the episode, the rendered \textit{Visual Sketch} image produced for the immediately preceding sub-task.
    
\item \textit{\textbf{Reasoning Mode Outputs:}}
    Triggered by the \texttt{<BOR>} (Begin-of-Reasoning) token, the LLM auto-regressively generates a textual spatial and temporal rationale followed by the \textit{Sub-Task} and the structured definition of the \textit{Visual Sketch}, including coordinates for bounding boxes, keypoints, and arrows (\textit{i.e.,} \texttt{\{"bbox": [...], "points": [...], "arrow": [...] \}}). This mode is terminated by \texttt{<EOR>} and texture  \textit{Visual Sketch} is then rendered into an image.

\item \textit{\textbf{Action Mode Inputs:}}
    Once the \textit{Sub-Task} and \textit{Visual Sketch} is generated and rendered, the model inputs are updated to include:
    (1) The current observation images $I^{1:n}_t$;
    (2) The rendered \textit{Visual Sketch} image, which explicitly overlays the spatial intent on the reference view;
    (3) The robot's current proprioceptive state $s_t$ (\textit{e.g.,} joint angles or end-effector pose).
    
\item \textit{\textbf{Action Mode Outputs:}}
    Once triggered by the \texttt{<BOA>} (Begin-of-Action) token from LLM, the Action Expert predicts a sequence of future actions (an action chunk) $A_t$. Specifically, it predicts the flow vector field to denoise a random noise sample into a valid trajectory of end-effector poses/joint angles and gripper states.
\end{itemize}

\begin{table*}[t]
\centering
    \caption{\textbf{Detailed configuration for each training stage of Action-Sketcher.} The table presents the training statistics and hyperparameters across the three-stage curriculum. Stage 1 focuses on foundational spatiotemporal representations, Stage 2 on reasoning-to-sketch generation, and Stage 3 on action policy alignment and mode adaptation.}
    \label{tab:training_setting}
    \setlength{\tabcolsep}{8pt}
    \renewcommand{\arraystretch}{1.3}
    \resizebox{0.93\textwidth}{!}{%
    \begin{tabular}{ll|c|c|c}
        \toprule
        & & \multicolumn{1}{c|}{\textbf{Stage 1}} & \multicolumn{1}{c|}{\textbf{Stage 2}} & \multicolumn{1}{c}{\textbf{Stage 3}} \\ 
        & & \small{(Foundational Spatiotemporal Learning)} & \small{(Reasoning-to-Sketch)} & \small{(Sketch-to-Action / Mode Adaptation)} \\
        \midrule 
        \multirow{3}{*}{\rotatebox[origin=c]{90}{\small \textit{Data}}}
        & \textbf{Dataset Source} & Grounding / Pointing / Planning & Reasoning-to-Sketch Dataset & Reasoning-Sketch-Action Corpus \\
        & \textbf{\#Samples} & $\sim$3.4M (Spatial) + 870k (Temp.) & $\sim$21k Samples & $\sim$6k Episodes (2.8M Samples) \\
        & \textbf{Data Modality} & Image + Text & Image + Text $\to$ Sketch & Image + Text + Sketch $\to$ Action \\
        \midrule 
        \multirow{3}{*}{\rotatebox[origin=c]{90}{\small \textit{Model}}}
        & \textbf{Base Architecture} & $\pi_0$ (PaliGemma-3B) & $\pi_0$ (PaliGemma-3B) & $\pi_0$ (PaliGemma-3B) \\
        & \textbf{Trainable Parts} & Vision Encoder + LLM & Vision Encoder + LLM & LLM + Action Expert \\
        & \textbf{Action Head} & Frozen & Frozen & Flow Matching \\
        \midrule 
        \multirow{12}{*}{\rotatebox[origin=c]{90}{\small \textit{Training}}}
        & \textbf{Global Batch Size} & 256 & 32 & 32 \\
        & \textbf{Gradient Accumulation} & 4 & 2 & 2 \\   
        & \textbf{LR: $\{\psi_v^{\text{ViT}}, \phi_v^{\text{LLM}}, \phi_v^{\text{AE}}\}$} & $\{1 \times 10^{-6},1 \times 10^{-4},N/A$\} & $\{1 \times 10^{-6},5 \times 10^{-5},N/A$\} & $\{N/A,1 \times 10^{-5},1 \times 10^{-4}\}$ \\
        & \textbf{LR Schedule} & Cosine Decay & Cosine Decay & Constant with Warmup \\
        & \textbf{Warmup Ratio} & 0.03 & 0.03 & 0.05 \\
        & \textbf{Optimizer} & AdamW & AdamW & AdamW \\
        & \textbf{Weight Decay} & 0.05 & 0.05 & 0.05 \\
        & \textbf{Image Resolution} & $224 \times 224$ & $224 \times 224$ & $224 \times 224$ \\
        & \textbf{Max Seq. Length} & 4096 & 8192 & 8192 \\
        & \textbf{Precision} & bfloat16 & bfloat16 & bfloat16 \\
        & \textbf{Training Epochs} & 1 & 1 & 20 \\
        & \textbf{Hardware} & 32 $\times$ H100 & 8 $\times$ H100 & 8 $\times$ H100 \\
        \bottomrule
    \end{tabular}
    }
    \vspace{-1.0em}
\end{table*}

\subsection{Training Setting}
In the main text of the paper, we employed a staged training strategy, with complete settings presented in Tab.~\ref{tab:training_setting}. During the entire training phase, we conducted all experiments on a cluster of servers, each equipped with 8$\times$H100 GPUs. We implement our models using PyTorch $\pi_0$ codebase. We utilize the AdamW optimizer with a cosine decay learning rate schedule and a weight decay of 0.05 across all stages. To ensure training stability and efficiency with high-resolution inputs ($224 \times 224$), we employ bfloat16 mixed-precision training. The specific curriculum details are as follows:
\begin{itemize}
\item\textbf{Stage 1: Foundational Spatiotemporal Learning.} 
To handle the large-scale pre-training corpus ($\sim$4.3M samples combined), we scale the training to 32 H100 GPUs. We use a global batch size of 256 to efficiently digest the diverse visual grounding and temporal planning data. In this stage, the Action Head is not initialized; we focus on tuning the Vision Encoder (LR $1\times 10^{-6}$) and the LLM backbone (LR $1\times 10^{-4}$) to master spatiotemporal reasoning within a context length of 4096 tokens.

\item\textbf{Stage 2: Reasoning-to-Sketch Alignment.} 
We fine-tune the model on the curated 21k-sample dataset to specialize in generating \textit{Visual Sketches}. The context window is expanded to 8192 tokens to accommodate long-horizon reasoning history and detailed sketch definitions. We reduce the global batch size to 32 and lower the LLM learning rate to $5\times 10^{-5}$ to preserve the pre-trained knowledge while aligning the text output with the specific JSON-based sketch format.

\item\textbf{Stage 3: Sketch-to-Action and Mode Adaptation.} 
In the final stage, we jointly train the full model, including the previously frozen components and the Flow Matching action head. To support the complex multimodal mapping from sketches to continuous actions, we employ a differential learning rate strategy: the Vision Encoder remains frozen to maintain feature stability, the LLM is fine-tuned with a reduced learning rate ($1 \times 10^{-5}$), and the Action Expert is trained with a higher rate ($1 \times 10^{-4}$). This stage runs for 20 epochs to ensure the action policy fully converges with the sketch-conditioned inputs.

\end{itemize}

\subsection{Evaluation Details}
\label{sec:eval_details}
We provide comprehensive details about our evaluation protocols, metrics, and experimental setup across both simulation and real-world environments. For real-world experiments, we deploy our trained policies on two robotic platforms: Aloha AgileX and Galaxea R1. 

\begin{itemize}

\item \textbf{RoboTwin 2.0 (Sim).} We evaluate on five custom manipulation tasks in the RoboTwin 2.0 simulation environment. Each task is evaluated over 100 rollouts using the same random seed across all models to ensure fair comparison. Success is determined by the simulator's built-in success condition, which checks whether the episode meets the task-specific completion criteria.

\item \textbf{Long-Horizon Tasks (Real).} For the two long-horizon manipulation tasks (Tidy Table and Pour Tea), we evaluate on 10 rollouts per model. To test robustness, we modify the scene every 2 rollouts by changing object positions and introducing new object types. All models are evaluated on the same scenario before transitioning to the next scene configuration.

\item \textbf{General Pick and Place (Real).} We evaluate on 30 rollouts per model, modifying the scenario every 5 rollouts to test generalization across diverse spatial arrangements and distractor objects. As with the long-horizon tasks, all models complete evaluation on each scenario before scene changes.
\end{itemize}

\section{Details of Training Dataset}
\label{sec2}
\begin{figure}[t]
    \centering
    \includegraphics[width=0.92\linewidth]{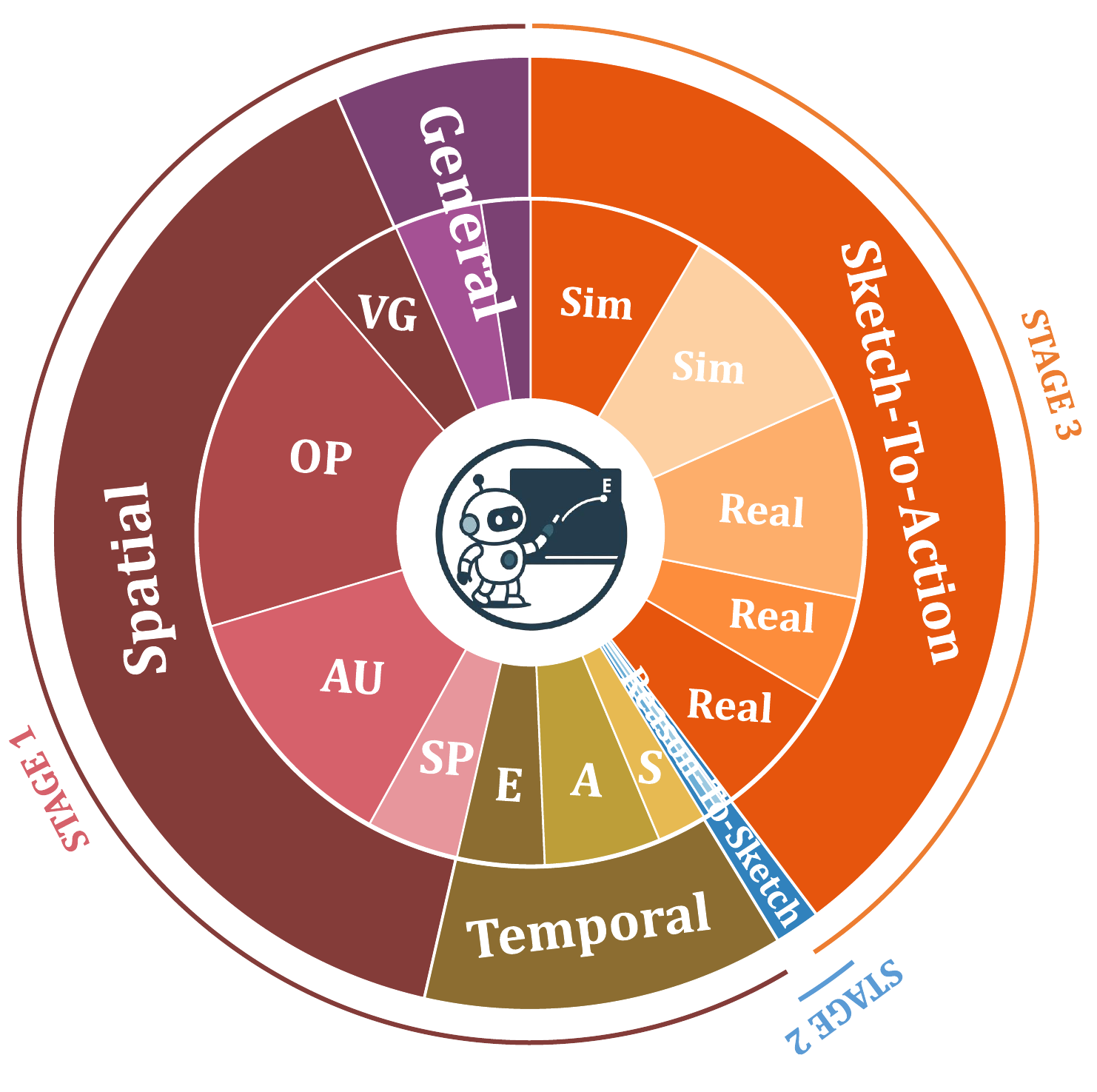}
    \caption{The distribution of the entire training dataset.}
    \label{fig:training_data}
\end{figure}

In the main body of the paper, we emphasize the importance of the training data and the proportion of robotic data. In this section, we will provide a detailed overview of the training data and its sources. The distribution of the entire training dataset is illustrated in Fig.~\ref{fig:training_data}.

\begin{itemize}
\item \textbf{General MLLM VQA (470k).} To establish a robust foundation for multimodal understanding, we construct a general visual question-answering dataset comprising 470k high-quality samples. This subset is primarily derived from two sets: LLaVA-665k~\cite{liu2024improved} and LRV-400k~\cite{liu2023mitigating}, encompassing diverse tasks such as standard VQA, OCR-based queries, and visual dialogues.

\item \textbf{Visual Grounding (327k).} We enhance object-level localization capabilities by leveraging extensive annotations from LVIS~\cite{gupta2019lvis}, Ref-L4~\cite{chen2024revisiting}, OV-VG~\cite{li2024llavaov, an2025llava}, RefCOCO~\cite{yu2016modeling,mao2016generation}, and Visual Genome~\cite{krishna2017visual}. In this subset, we include 327k high-resolution images with the standardized bounding box coordinates \((x_1,y_1,x_2,y_2)\). 

\item \textbf{Object Pointing (1.3M).} This module focuses on precise coordinate identification in embodied contexts. We process the Pixmo-Points~\cite{deitke2024molmo} dataset through a two-step filtering mechanism, \textit{i.e.,} limiting point counts and utilizing GPT-4o~\cite{hurst2024gpt4o} to select indoor-relevant scenes, resulting in 190k QA pairs optimized for reduced clutter. Additionally, we incorporate 347k object reference samples from RoboPoint~\cite{yuan2024robopoint} and extend capabilities to physical-world interactions using the Spatial Referring Dataset~\cite{zhou2025roborefer} (802k samples), ensuring the model can generalize across diverse pointing tasks.

\item \textbf{Affordance Understanding (881k).} We address both functional and spatial affordances to support object interaction. For functional reasoning, we utilize part-level annotations from PACO-LVIS~\cite{ramanathan2023paco} to generate 561k QA pairs regarding object functionality and part usage. For spatial affordance, we integrate 320k region-reference pairs from RoboPoint, where coordinate annotations are optimized to help the model identify valid placement areas and spatial relationships in real-world settings.

\item \textbf{Spatial Perception (318k).} To enable fine-grained 3D environmental understanding, we integrate multiple 3D-centric datasets: MMScan-224k~\cite{mmscan} for segmentation and description, 3RScan-43k~\cite{3rscan} for semantic reconstruction, as well as ScanQA-25k~\cite{scanqa} and SQA3D-26k~\cite{sqa3d} for spatially grounded question answering.

\item \textbf{Temporal Planning (870k).} To empower the model with robust temporal reasoning and execution capabilities, we construct a unified planning dataset comprising 870k samples. This collection synthesizes diverse embodied scenarios from three primary sources: EgoPlan-IT~\cite{chen2023egoplan} for egocentric task planning, ShareRobot~\cite{ji2025robobrain} to support fine-grained manipulation, and AgiBot-World~\cite{bu2025agibot} for continuous visual sequences from diverse household environments. This subset facilitates long-horizon task decomposition and sequential reasoning.

\item \textbf{Reasoning-To-Visual (21k).} To bridge high-level semantic reasoning with precise visual grounding, we construct a specialized dataset containing 21k high-quality instruction-to-sketch pairs. Data generation employs a dual-pipeline strategy: an automatic pipeline leveraging SAM~\cite{kirillov2023sam} and GPT-4o for scalable annotation, and a human-in-the-loop pipeline utilizing a custom interface to ensure geometric accuracy. Each sample pairs a textual sub-task instruction with a structured Visual-Sketch (bounding boxes, keypoints, and directional arrows), augmented by GPT-4o-generated reasoning chains that justify both the temporal necessity of the action and the spatial rationale behind the visual cues.

\item \textbf{Visual-To-Action (2.8M).} To translate visual plans into executable motor controls, we compile a large-scale trajectory dataset aligning Visual-Sketches with low-level action sequences. This corpus spans 6.1k episodes, resulting in 2.8M state-action pairs derived from LIBERO~\cite{liu2023libero}, a modified complex version of RoboTwin 2.0~\cite{chen2025robotwin}, and real-world teleoperation data. Each sample integrates multi-view observations, the generated Visual-Sketch, and reasoning context to predict precise action chunks, ensuring the model learns to ground its manipulation policy effectively in the provided visual guidance.

\end{itemize}

\begin{table*}[t]
\centering
\caption{RoboTwin 2.0 and Real-world task specifications.}
\label{tab:all_tasks}
\setlength{\tabcolsep}{4pt}
\renewcommand{\arraystretch}{1.15}
\resizebox{1.8\columnwidth}{!}{%
\begin{tabular}{llcc}
\toprule
\textbf{Task} & \textbf{Description} & \textbf{Robot} & \textbf{\#Episodes} \\
\midrule
\rowcolor[HTML]{F2F2F2} \multicolumn{4}{l}{\textbf{\textit{RoboTwin 2.0 Tasks}}} \\
\midrule
Stack Blocks   & Stack three blocks with varying color orders (RGB/GBR) & AgileX Piper & 550 \\
Hanging Mug    & Grasp and hang mug on rack & AgileX Piper & 200 \\
Place Cups     & Place two cups to nearest plates among randomly positioned plates & AgileX Piper & 200 \\
Place A2B      & Place object relative to another's left or right side & AgileX Piper & 550 \\
\midrule
\rowcolor[HTML]{F2F2F2} \multicolumn{4}{l}{\textbf{\textit{Real-world Tasks}}} \\
\midrule
Tidy Table     & Place diverse objects in containers (16 sub-tasks) & AgileX Piper   & 342 \\
Pour Tea       & Pour tea into two teacups (7--8 sub-tasks)         & AgileX Piper   & 749 \\
Pick \& Place  & Place object amid distractors                      & Galaxea R1 & 1,500 \\
\bottomrule
\end{tabular}}
\end{table*}

\section{Details of Data Construction}
\label{sec2_ext}

\noindent For Stage 2, to equip the model with downstream task-reasoning proficiency and the capability to generate semantically grounded Visual Sketches, we instantiate three independent annotation pipelines: (i) an automatic Visual-Sketch pipeline, (ii) a human Visual-Sketch pipeline, and (iii) an automatic Reasoning-generation pipeline that is grounded on the Visual-Sketch data produced by the first two pipelines.  
This stage is correspondingly organized into two complementary subsections: Visual-Sketch generation (Sec.~\ref{vsg}) and Reasoning generation (Sec.~\ref{rg}). For Stage 3, to construct the full Reasoning-Sketch-Action training corpus, we integrate the Visual-Sketches and reasoning annotations from Stage 2 with the corresponding action sequences in the demonstration trajectories (Sec.~\ref{action}).

\subsection{Visual-Sketch Generation}
\label{vsg}
\begin{figure}[t]
    \centering
    \includegraphics[width=0.95\linewidth]{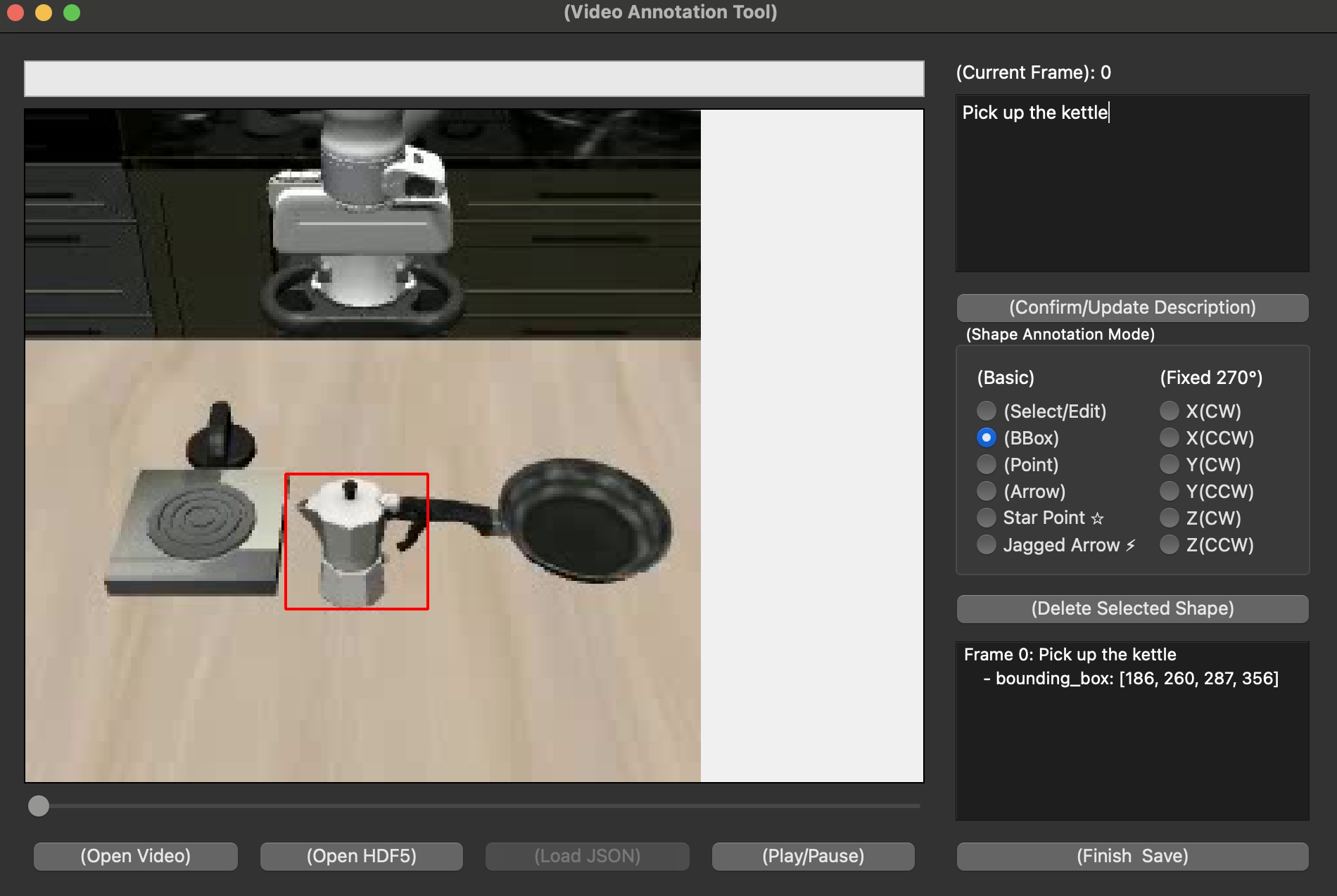}
    \caption{\textbf{Human annotation interface.} Screenshot of our PyQt-based annotation tool showing the interface for creating Visual-Sketch annotations. Annotators can navigate frames, overlay bounding boxes, keypoints, and directional arrows, and edit sub-task descriptions in real-time.}
    \label{fig:annotation_tool}
\end{figure}

\begin{itemize}
\item \textbf{Automatic annotation pipeline.}  
Each demonstration is temporally decomposed into sub-tasks by detecting consecutive gripper open/close transitions.  SAM~\cite{kirillov2023sam} is applied to the start and end frames of every segment to produce instance masks; the target object is designated as the mask exhibiting maximal centroid displacement.  The egocentric image pair, the target's bounding box, and its 2-D start and end coordinates are forwarded to GPT-4o with a deterministic prompt that requests (i) a concise sub-task instruction and (ii) a JSON-compliant Visual-Sketch record containing bounding box, keypoints, and arrows.
\item \textbf{Human annotation pipeline.}  
We implement a PyQt interface (Fig.~\ref{fig:annotation_tool}) that streams HDF5, MP4, JPG, or PNG sequences.  Annotators navigate to any frame, overlay the same JSON primitives, and validate or rewrite the sub-task description; the interface enforces schema consistency and averages 2.3 subtask annotations per minute.  The identical tool is optionally deployed for human-in-the-loop refinement during inference, allowing real-time correction of model-generated sketches.
\end{itemize}

\smallskip
\noindent
\textbf{Quality assessment.}  
Ablations in Sec.~\ref{sec3} reveal that both pipelines improve success rates over baseline in simulation, but the human-curated set yields a further +7.6\% absolute gain, primarily ascribed to fewer inaccurate sketches.  Consequently, all main-results models are trained on the human-refined split, while the automatic pipeline will be released as an open-source tool to enable scalable future expansion.
\subsection{Reasoning Generation}
\label{rg}
\begin{figure*}[t]
    \centering
    \includegraphics[width=0.96\linewidth]{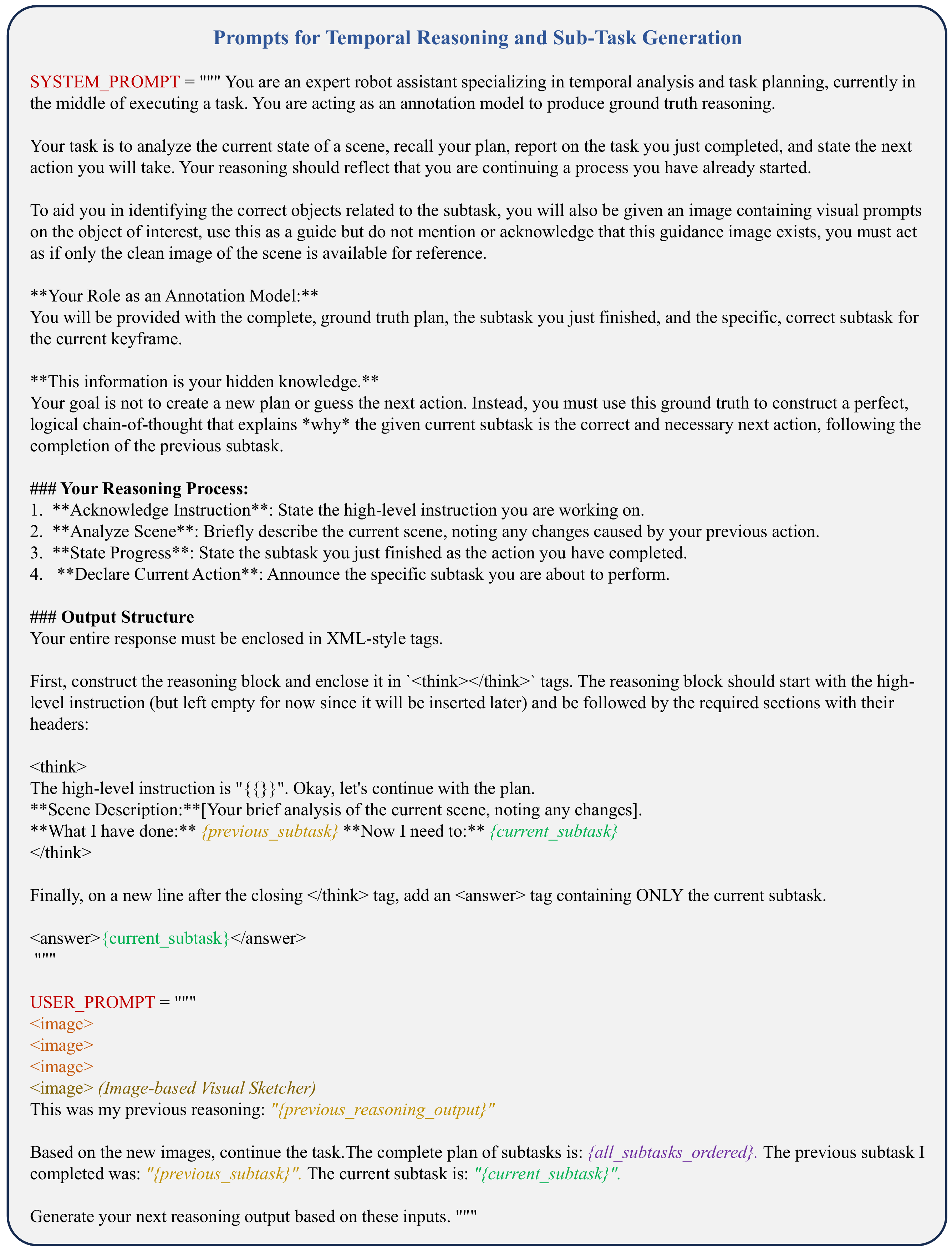}
    \caption{Temporal Reasoning Prompt.}
    \label{fig:prompt_tmp}
\end{figure*}

\begin{figure*}[t]
    \centering
    \includegraphics[width=0.96\linewidth]{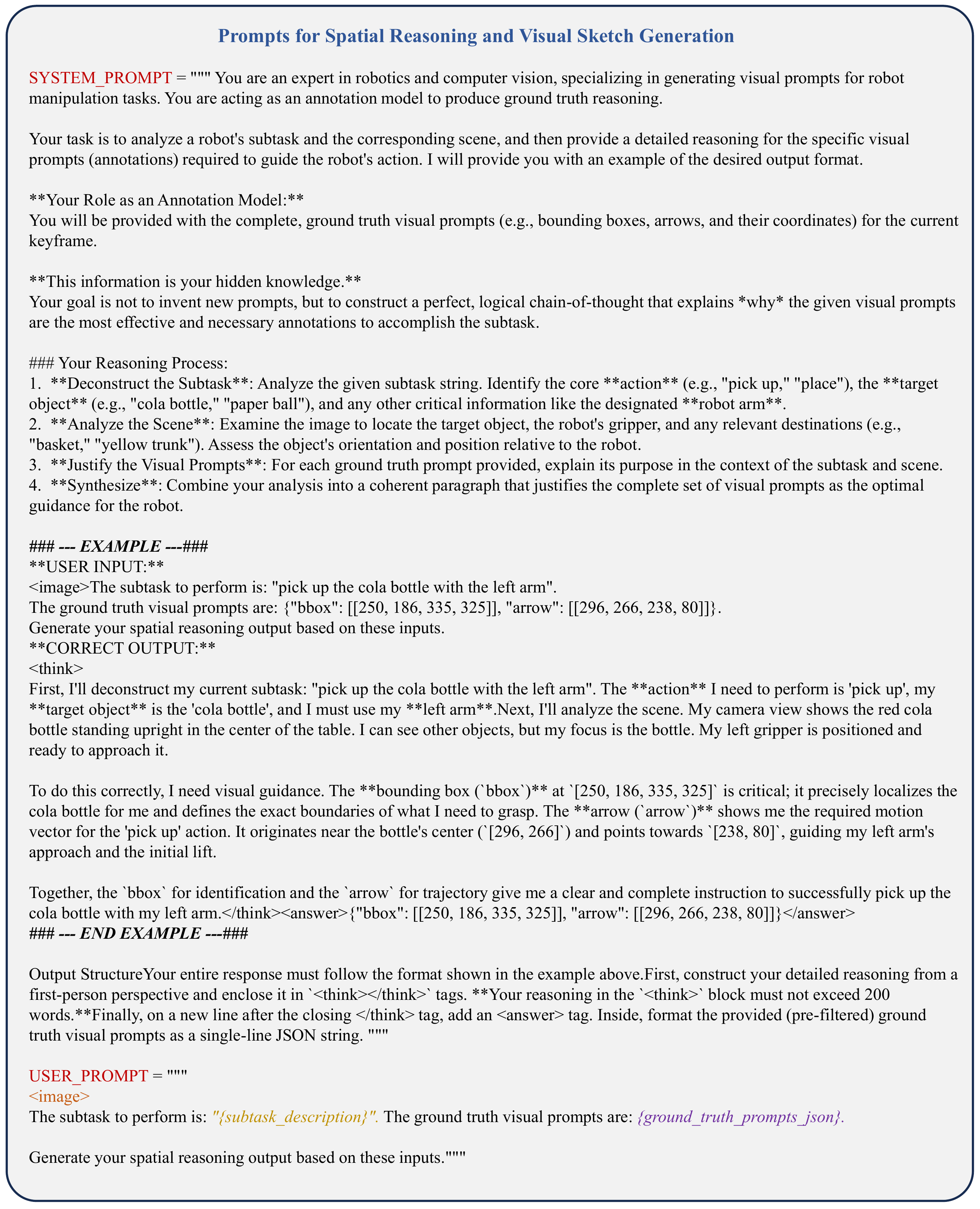}
    \caption{Spatial Reasoning Prompt.}
    \label{fig:prompt_spa}
\end{figure*}

Reasoning data are produced along two complementary axes: temporal and spatial.
Both branches leverage GPT-4o conditioned on the Visual-Sketch annotations.
\begin{itemize}
\item \textbf{Temporal reasoning.}  
Given an episode-level list of sub-tasks and the Visual-Sketch image for the current key-frame, we prompt GPT-4o to produce a concise chain-of-thought that (i) recalls the high-level instruction, (ii) summarises the scene state, (iii) states the previously completed sub-task, and (iv) justifies the current sub-task as the next necessary action. The complete prompt template is provided in Fig.~\ref{fig:prompt_tmp}.

\item \textbf{Spatial reasoning.}  
For every key-frame we additionally query GPT-4o to explain \emph{why} the ground-truth visual prompts are the minimal set required to accomplish the current sub-task. The prompt conditions on the sub-task string and the JSON-encoded sketch; the complete template is provided in Fig.~\ref{fig:prompt_spa}.
\end{itemize}

\subsection{Sketch-Action Generation}
\label{action}
For each episode, we align the generated sub-tasks and Visual-Sketches with the corresponding action sequences. Crucially, all action samples belonging to the same sub-task share the same reference Visual-Sketch, ensuring consistent spatial grounding throughout sub-task execution. This results in approximately 2.8M action prediction samples across 6.1k episodes, with each sample containing multi-view observations, the rendered Visual-Sketch for the current sub-task, reasoning text, and ground-truth action chunks.
Our training corpus combines simulation and real-world data to ensure robust generalization:

\begin{itemize}
\item \textbf{LIBERO~\cite{liu2023libero}.} We utilize the open-sourced training sets from four task suites: LIBERO-Spatial, LIBERO-Object, LIBERO-Goal, and LIBERO-Long, which provide diverse manipulation scenarios with varying spatial reasoning and long-horizon planning requirements. We use third-person and wrist camera images along with robot proprioceptive state and language instructions.

\item \textbf{RoboTwin 2.0~\cite{chen2025robotwin}.} 
We select 5 tasks from RoboTwin 2.0 and introduce modifications to increase complexity, including random object clutter, increased camera height to reduce object size, and extended episode length. These modifications ensure tasks are sufficiently long ($\geq$4 sub-tasks) and test spatial understanding. Task details are summarized in Tab.~\ref{tab:all_tasks}. All tasks use the Aloha Agile-X embodiment with egocentric and left/right wrist views, controlled via absolute joint positions.

\item \textbf{Real-World Data.}
To validate real-world transfer, we collect two long-horizon manipulation tasks and one spatial understanding task in laboratory settings. Details are provided in Tab.~\ref{tab:all_tasks}. All tasks use absolute joint position control with egocentric and wrist camera views.
\end{itemize}

\section{Complementary Experiments}
\label{sec3}
In this section, we present the complete experiments and results that are omitted from the main manuscript due to page limitations.
We present supplementary experiments that provide further insights into our approach but were omitted from the main manuscript due to space constraints.

\subsection{Ablation Experiments in Detail}
In the main manuscript, we conducted several ablation studies to validate key design choices of Action-Sketcher. Here we provide additional details and deeper analysis of those experiments. To validate the necessity of each component in our Action-Sketcher framework, we conducted controlled ablations by removing or modifying specific modules while keeping all other training and evaluation protocols identical.

\textbf{w/o Spatial Reasoning.} We trained a model to directly generate Visual-Sketches without the intermediate spatial reasoning step. Specifically, the model predicts the JSON-encoded sketch parameters directly after temporal reasoning, bypassing the chain-of-thought explanation about why specific visual prompts are necessary. This ablation isolates the impact of explicit spatial reasoning on generating accurately grounded Visual-Sketches.

\textbf{w/o Visual Sketches.} We trained a model that undergoes the same reasoning steps as the original Action-Sketcher (generating both temporal and spatial reasoning), but Visual-Sketches are not rendered or overlaid on the reference images during action execution. Instead, only the original observation images are provided to the action prediction module. This ablation isolates the impact of using Visual-Sketches as explicit spatial reference during action execution, as opposed to simply providing additional reference images.

The results demonstrate that both spatial reasoning and Visual-Sketch conditioning contribute significantly to task performance, with the full Action-Sketcher framework achieving the best results across all evaluation metrics.

\subsection{Sub-task vs.\ Task-level Visual Sketching}
Our first supplementary experiment studies the effect of the granularity of visual sketching. We compare task-level sketches, where a single Visual-Sketch is provided for the entire task, against sub-task-level sketches, where a separate Visual-Sketch is given for each annotated sub-task (Action-Sketcher). Both variants are evaluated on the same RoboTwin~2.0 tasks in simulation, and use identical training and evaluation protocols except for the sketching level. The results are reported in Tab.~\ref{tab:subtask_vs_task_sketch}.

\begin{table}[h]
\centering
\caption{Sub-task vs task-level visual sketching on RoboTwin 2.0 simulation tasks (success rate \%).}
\label{tab:subtask_vs_task_sketch}
\small
\setlength{\tabcolsep}{3pt}
\resizebox{0.46\textwidth}{!}{%
\begin{tabular}{lccccc|c}
\toprule
\textbf{Method} & Stack & Hang & Empty & A2B-L & A2B-R & \textbf{Avg.} \\
\midrule
Task-Level & 9.5 & 20.0 & 15.0 & 35.5 & 21.0 & 20.2 \\
\textbf{Sub-Task (Ours)} & \textbf{34.5} & \textbf{25.0} & \textbf{28.0} & \textbf{43.0} & \textbf{28.0} & \textbf{31.7} \\
\midrule
$\Delta$ Improvement & +25.0 & +5.0 & +13.0 & +7.5 & +7.0 & +11.5 \\
\bottomrule
\end{tabular}}
\end{table}

\noindent\textbf{Analysis.} The results reveal a clear advantage of sub-task-level sketching across all tasks, with our approach achieving an average improvement of 11.5\% absolute over task-level sketching. Notably, the performance gap varies significantly based on task complexity and length. The Stack Blocks task, which requires precise sequential manipulation across 6 sub-tasks with strict spatial ordering constraints, shows the largest improvement (+25.0\%), while the redundant Hanging Mug task shows a smaller but still substantial gain (+5.0\%).

We hypothesize that as tasks grow longer with more sub-tasks, reasoning about the entire trajectory at once and generating a single comprehensive Visual-Sketch actually introduces more noise and hurts performance. Task-level sketching requires the model to simultaneously plan multiple manipulation stages, leading to less precise spatial grounding for individual actions and potential conflicts between overlapping visual prompts. In contrast, our sub-task-level approach decomposes complex tasks into manageable reasoning steps, allowing the model to focus on generating accurate Visual-Sketches for each immediate manipulation goal. This demonstrates the advantage of our method not only in general performance but especially on longer, more complex tasks where decomposed reasoning and progressive sketch generation are critical for success.

\subsection{Automated vs.\ Human Annotation}
The second supplementary experiment compares the performance of policies trained with human-annotated Visual-Sketches to those trained with sketches produced fully automatically by our pipeline. The human-annotated pipeline consistently achieves higher success rates, while our automated pipeline still performs better than the baseline. Results are summarised in Tab.~\ref{tab:human_vs_auto_annotation}.

\begin{table}[h]
\centering
\caption{Human vs automated annotation on RoboTwin 2.0 simulation tasks (success rate \%).}
\label{tab:human_vs_auto_annotation}
\small
\setlength{\tabcolsep}{3pt}
\resizebox{0.46\textwidth}{!}{%
\begin{tabular}{lccccc|c}
\toprule
\textbf{Method} & Stack & Hang & Empty & A2B-L & A2B-R & \textbf{Avg.}\\
\midrule
Automated & 21.5 & 21.0 & 23.0 & 32.0 & 23.0 & 24.1\\
\textbf{Human} & \textbf{34.5} & \textbf{25.0} & \textbf{28.0} & \textbf{43.0} & \textbf{28.0} & \textbf{31.7}\\
\midrule
$\Delta$ Improvement & +13.0 & +4.0 & +5.0 & +11.0 & +5.0 & +7.6 \\
\bottomrule
\end{tabular}}
\end{table}

\subsection{Inference Speed Analysis}
To evaluate the computational efficiency of Action-Sketcher, we analyze inference speed across different components and compare with baseline models. We measure Visual-Sketch generation time and episode execution time on representative tasks from simulation and real-world environments. All measurements are conducted on a single NVIDIA RTX 4090 GPU for both simulation and real-world evaluations. For each task, we calculate the average time to reach sub-task frames and the time to complete reasoning and sketch generation, averaged over 5 episodes. The results are reported in Tab.~\ref{tab:inference_speed}.

\begin{table}[h]
\centering
\caption{Inference speed and success rate analysis. Sketch time measures reasoning/sketch generation overhead per sub-task, Action time is average episode execution time, SR/Time is success rate normalized by total time ($\times 10^{-3}$).}
\label{tab:inference_speed}
\footnotesize
\setlength{\tabcolsep}{3pt}
\renewcommand{\arraystretch}{1.15}
\resizebox{0.85\columnwidth}{!}{%
\begin{tabular}{l|cc|cc|cc}
\toprule
& \multicolumn{2}{c|}{\textbf{Stack Blocks}} & \multicolumn{2}{c|}{\textbf{Hang Mug}} & \multicolumn{2}{c}{\textbf{Tidy Table}} \\
\cmidrule(lr){2-3} \cmidrule(lr){4-5} \cmidrule(lr){6-7}
\textbf{Metric} & $\pi_0$ & \textbf{Ours} & $\pi_0$ & \textbf{Ours} & $\pi_0$ & \textbf{Ours} \\
\midrule
Sketch (s) & -- & 3.8 & -- & 3.5 & -- & 6.4 \\
Action (s) & 9.0 & 9.4 & 13.1 & 13.0 & 26.0 & 24.5 \\
\midrule
Total Time (s) & 9.0 & 13.2 & 13.1 & 16.5 & 26.0 & 30.9 \\
SR (\%) & 4.0 & \textbf{34.5} & 20.0 & \textbf{25.0} & 23.0 & \textbf{52.0} \\
\midrule
SR/Time & 0.44 & \textbf{2.61} & \textbf{1.53} & 1.52 & 0.88 & \textbf{1.68} \\
\bottomrule
\end{tabular}%
}
\end{table}

\begin{table*}[!ht]
    \centering
    \caption{\textbf{Performance across five spatial reasoning benchmarks.} The best results among different models are highlighted in \textbf{bold}, while the second-best results are \underline{underlined}.}
    \vspace{-0.5em}
    \resizebox{0.95\textwidth}{!}{
    \begin{tabular}{l|ccc|c|c|c|ccc}
        \toprule
        \multicolumn{1}{l|}{\multirow{2}{*}{\textbf{Models / Metrics}}}  & \multicolumn{3}{c|}{\textbf{BLINK}}  &  \textbf{CV-Bench} &  \textbf{EmbSpatial} &  \textbf{RoboSpatial} &  \multicolumn{3}{c}{\textbf{RefSpatial-Bench}}\\ 
         \cmidrule(lr){2-4} \cmidrule(lr){5-5} \cmidrule(lr){6-6} \cmidrule(lr){7-7} \cmidrule(lr){8-10}
        & \textbf{Dep.} & \textbf{Spa.} & \textbf{All $\uparrow$} & \textbf{All $\uparrow$} & \textbf{All $\uparrow$} & \textbf{All $\uparrow$} & \textbf{Loc.} & \textbf{Pla.} & \textbf{All $\uparrow$} \\ 
        \midrule
        \rowcolor[HTML]{F2F2F2} \multicolumn{10}{l}{\textbf{General Baselines}} \\ \midrule
        Gemini-2.5-Pro-preview-05-06~\cite{gemini25pro} & 79.03 & 84.62 & 81.83 & 84.59 & \textbf{78.74} & \underline{59.87} & \underline{44.58} & 31.73 & 38.16\\ 
        Gemini-2.5-Flash-preview-04-17~\cite{gemini25pro} & 77.42  & 79.02 & 78.22 & 84.03  & 74.75  & 54.10  & 37.50  & 23.00 & 30.25 \\
        GPT-o4-mini-2025-05-16~\cite{gpto3-o4-mini} & 79.03 & \textbf{88.11} & 83.57 & \underline{85.21} & 78.29 & 51.25 & 15.00 & 19.58 & 17.29\\ 
        GPT-4o-2024-11-20~\cite{gpt4o} & 72.58 & 83.22 & 77.90  & 78.63 & 71.92 & 44.42 & 8.00 & 9.55 & 8.78\\ 
        Claude-Sonnet-4-2025-05-14~\cite{claude4} & 75.81 & 80.42 & 78.12  & 78.43 & 64.26 & 51.26 & 5.00 & 10.37 & 7.69\\
        Qwen2.5-VL-32B-Instruct~\cite{Qwen2.5-VL} & 77.42 & 85.31 & 81.37 & 81.59 & 74.45 & 52.16 & 16.83 & 10.60 & 13.72\\ 
        Qwen2.5-VL-72B-Instruct~\cite{Qwen2.5-VL}& 74.19 & 78.32 & 76.26 & 82.68 & 73.30 & 48.33 & 23.50 & 15.83 & 19.67 \\ 
        \midrule
        \rowcolor[HTML]{F2F2F2} \multicolumn{10}{l}{\textbf{Embodied Baselines}} \\ \midrule
        Cosmos-Reason1-7B~\cite{cosmos-reason1}  & 63.71 & 73.43 & 68.57 & 74.71 & 65.22 & 38.81 & 9.84 & 1.04 & 5.44 \\ 
        VeBrain-8B~\cite{luo2025vebrain}  & 78.23 & 81.12 & 79.68 & 78.57 & 70.52 & 42.48 & 0.03 & 0.57 & 0.30 \\ 
        Magma-8B~\cite{magma} & 65.32  & 66.43 & 65.88 & 60.98  & 64.59  & 33.71  & 1.00  & 8.00 & 4.50 \\
        RoboBrain-7B-1.0~\cite{ji2025robobrain}  & 75.81 & 78.32 & 77.07 & 76.22 & 68.13 & 51.53 & 14.43 & 5.41 & 9.92 \\ 
        RoboBrain-7B-2.0  & \textbf{84.68} & 83.22 & \textbf{83.95} & \textbf{85.75}& 76.32 & 54.23 & 36.00 & 29.00 & 32.50 \\ 
        RoboBrain-32B-2.0 & \underline{79.84} & \underline{87.41} & \underline{83.63} & 83.92 & \underline{78.57} & \textbf{72.43} & \textbf{54.00} & \textbf{54.00} & \textbf{54.00} \\ 
        \textbf{Action-Sketcher-3B (Ours)} & 82.30 & 74.67 & 82.00 & 82.89 & 72.65 & \underline{62.28} & 44.50 & \underline{41.00} & \underline{42.50} \\
        \bottomrule
    \end{tabular}
    }
    \label{tab:Synthetic}
\end{table*}

\noindent\textbf{Analysis.} Despite the 3.5--6.4s overhead from Visual-Sketch generation per sub-task, Action-Sketcher maintains competitive inference efficiency. When normalized by total time, Action-Sketcher achieves superior or comparable efficiency (SR/Time) across all tasks, with dramatic improvements on complex tasks like Stack Blocks (5.9×) and Tidy Table (1.9×). This demonstrates that performance gains from explicit spatial reasoning outweigh the computational overhead.

\subsection{Performance on Embodied Benchmarks}
\label{sec:benchmark_performance}

To validate that the spatial reasoning capabilities developed through our training curriculum transfer beyond manipulation tasks, we evaluate Action-Sketcher on a diverse set of embodied reasoning benchmarks. These benchmarks assess various aspects of spatial understanding, including depth perception, spatial relationship reasoning, affordance understanding, and trajectory prediction.

\textbf{Evaluation Setup.}
We evaluate on five established benchmarks spanning both synthetic and real-world scenarios: BLINK~\cite{fu2024blink}, CV-Bench~\cite{tong2024cambrian}, EmbSpatial-Bench~\cite{du2024embspatial}, RoboSpatial~\cite{song2025robospatial}, and RefSpatial-Bench~\cite{zhou2025roborefer}. These benchmarks collectively evaluate spatial reasoning across diverse modalities, including visual question answering, coordinate prediction, affordance reasoning, and trajectory forecasting.

\textbf{Baseline Comparison.}
We compare Action-Sketcher against two categories of baselines: (1) \textit{\textbf{General VLMs}}: state-of-the-art vision-language models including Gemini-2.5-Pro~\cite{gemini25pro}, GPT-4o~\cite{gpt4o}, Claude-Sonnet-4~\cite{claude4}, and Qwen2.5-VL~\cite{Qwen2.5-VL}; (2) \textit{\textbf{Embodied Models}}: specialized models for robotic reasoning including Cosmos-Reason1~\cite{cosmos-reason1}, VeBrain~\cite{luo2025vebrain}, Magma~\cite{magma}, RoboBrain~\cite{ji2025robobrain, team2025robobrain}.

\textbf{Results and Analysis.}
Tab.~\ref{tab:Synthetic} presents comprehensive results across all benchmarks. Action-Sketcher demonstrates strong performance, achieving competitive or state-of-the-art results on multiple benchmarks. Notably, our model achieves particularly strong performance on benchmarks that emphasize spatial grounding and coordinate prediction (e.g., RoboSpatial, RefSpatial-Bench), validating that our Visual-Sketch training approach effectively develops fine-grained spatial reasoning capabilities. 

The results show that Action-Sketcher's explicit spatial reasoning training, where the model learns to generate grounded visual prompts with precise coordinates, transfers effectively to general spatial reasoning tasks beyond manipulation. This suggests that our framework not only improves manipulation performance but also develops more robust spatial understanding that generalizes across diverse embodied reasoning scenarios.

Here are the key findings:
\begin{itemize}
\item Action-Sketcher outperforms specialized embodied models on spatial benchmarks (RoboSpatial, RefSpatial-Bench), demonstrating the effectiveness of Visual-Sketch training for developing precise spatial reasoning.
\item Our model achieves competitive performance with significantly larger general VLMs while being specifically optimized for embodied tasks with much fewer parameters.
\item Performance improvements are most pronounced on tasks requiring coordinate-level spatial understanding, validating our design choice of explicitly training models to generate grounded visual prompts.
\end{itemize}

\section{More Qualitative Results}
\label{sec4}
In this section, we provide additional visualizations to demonstrate the effectiveness of Action-Sketcher's reasoning and visual sketch generation process across diverse manipulation tasks.

\subsection{Reasoning and Visual Sketch Generation}
\begin{figure*}[t]
    \centering
    \includegraphics[width=0.96\linewidth]{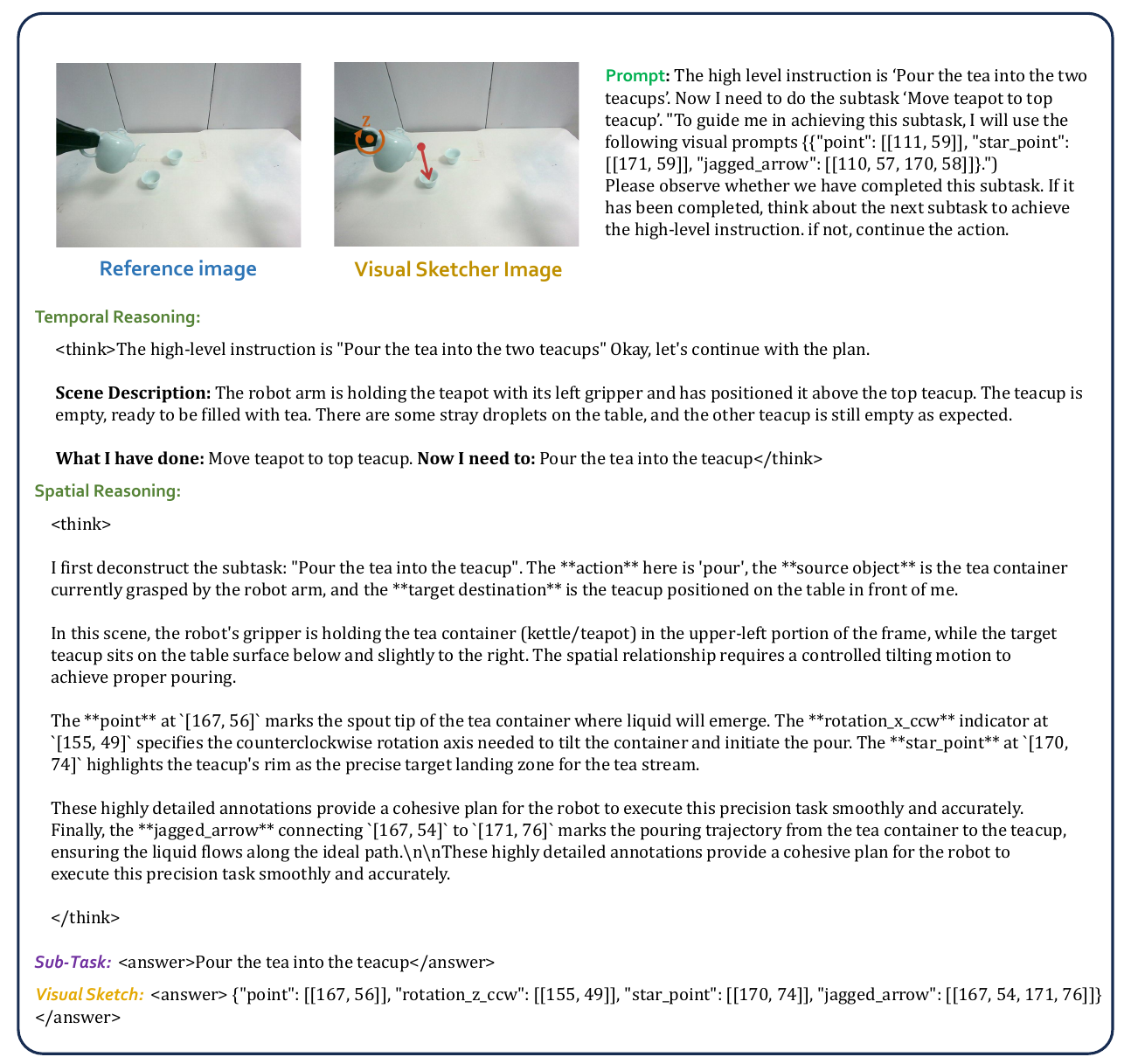}
    \caption{Visualization of Sub-Task and Visual-Sketcher Reasoning.}
    \label{fig:demo1}
\end{figure*}

\begin{figure*}[t]
    \centering
    \includegraphics[width=0.96\linewidth]{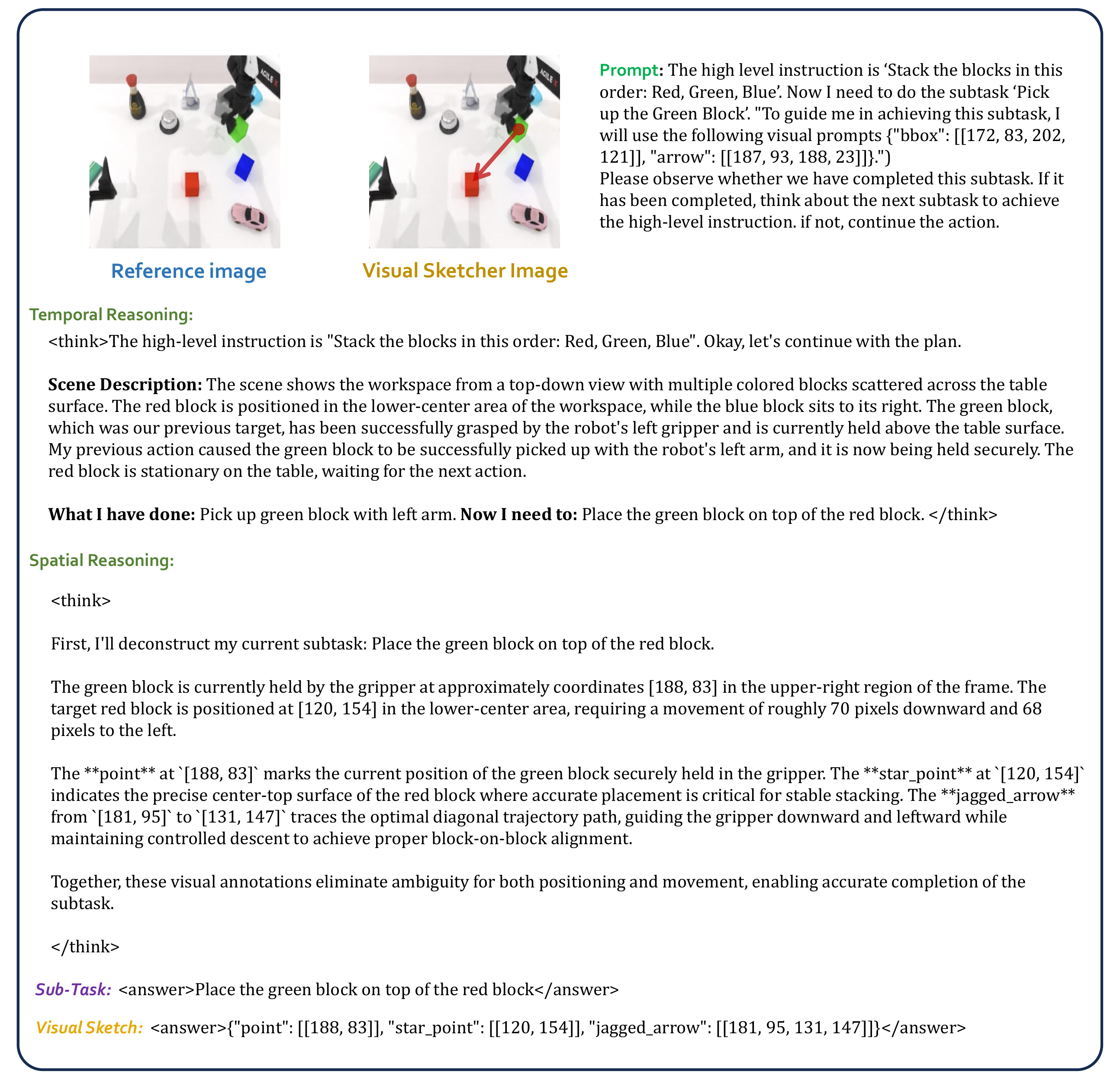}
    \caption{Visualization of Sub-Task and Visual-Sketcher Reasoning.}
    \label{fig:demo2}
\end{figure*}

\begin{figure*}[t]
    \centering
    \includegraphics[width=0.96\linewidth]{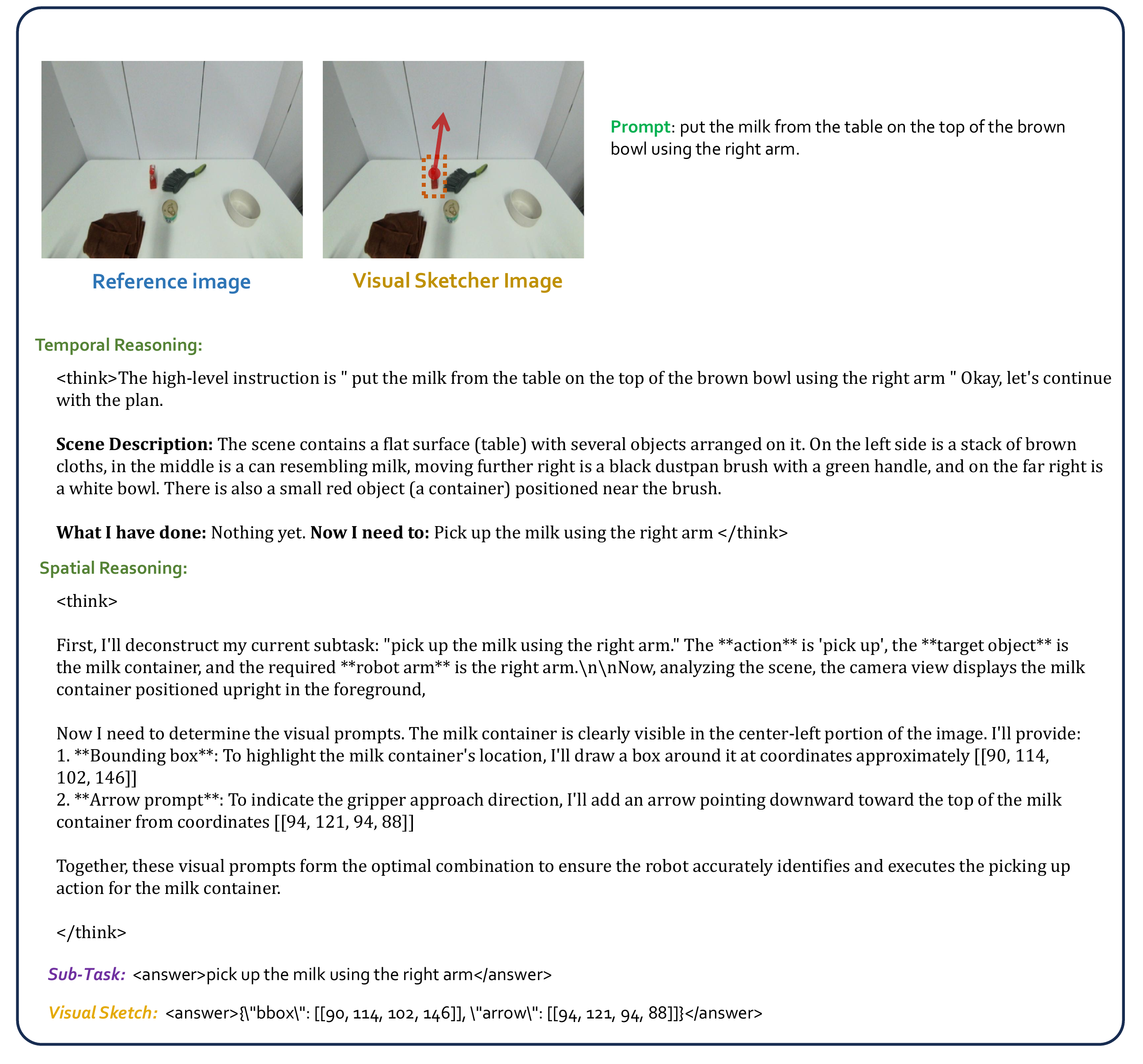}
    \caption{Visualization of Sub-Task and Visual-Sketcher Reasoning.}
    \label{fig:demo3}
\end{figure*}

\begin{figure*}[t]
    \centering
    \includegraphics[width=0.96\linewidth]{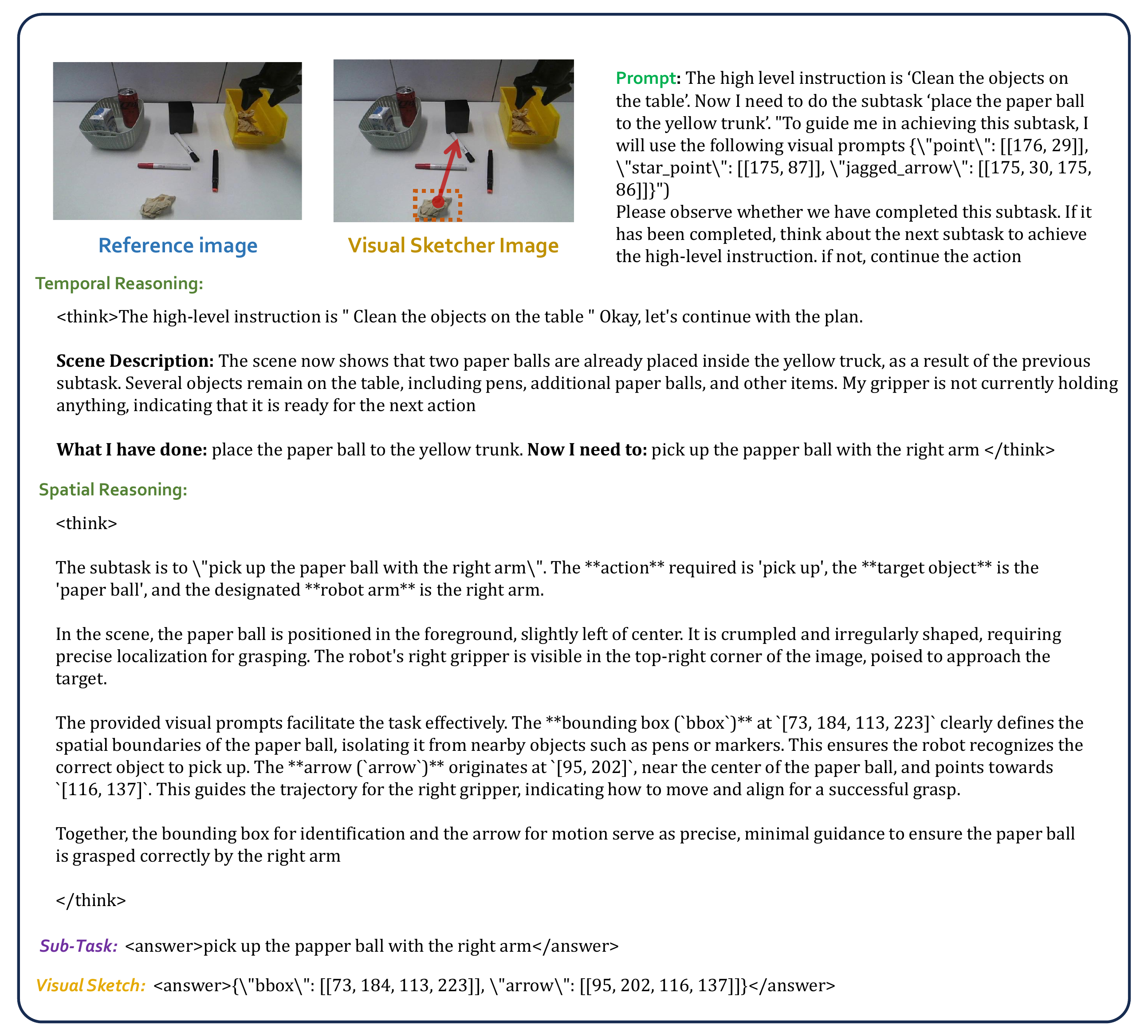}
    \caption{Visualization of Sub-Task and Visual-Sketcher Reasoning.}
    \label{fig:demo4}
\end{figure*}

Fig.~\ref{fig:demo1}-\ref{fig:demo4} illustrates the core reasoning part of complete See-Think-Sketch-Act cycle across multiple representative manipulation tasks. The visualizations demonstrate how Action-Sketcher decomposes high-level instructions into sub-tasks, generates spatial and temporal reasoning, produces grounded Visual-Sketches, and executes actions accordingly.
Each figures shows the reasoning process for a different task: (a) pour tea shows the specific spatial reasoning of subtasks, (b) block stacking demonstrating sequential manipulation with precise spatial ordering, (c) general object grasping showing spatial relationship understanding, (d) long-horizon manipulation illustrating multi-step task decomposition with contextual reasoning. For each step, the model generates detailed scene descriptions, reflects on completed actions, and explicitly reasons about the next necessary sub-task. The Visual-Sketches precisely encode spatial information through bounding boxes identifying target objects, arrows indicating motion directions, points marking current positions, and star points designating goal locations.

\begin{figure}[t]
    \centering
    \includegraphics[width=0.98\linewidth]{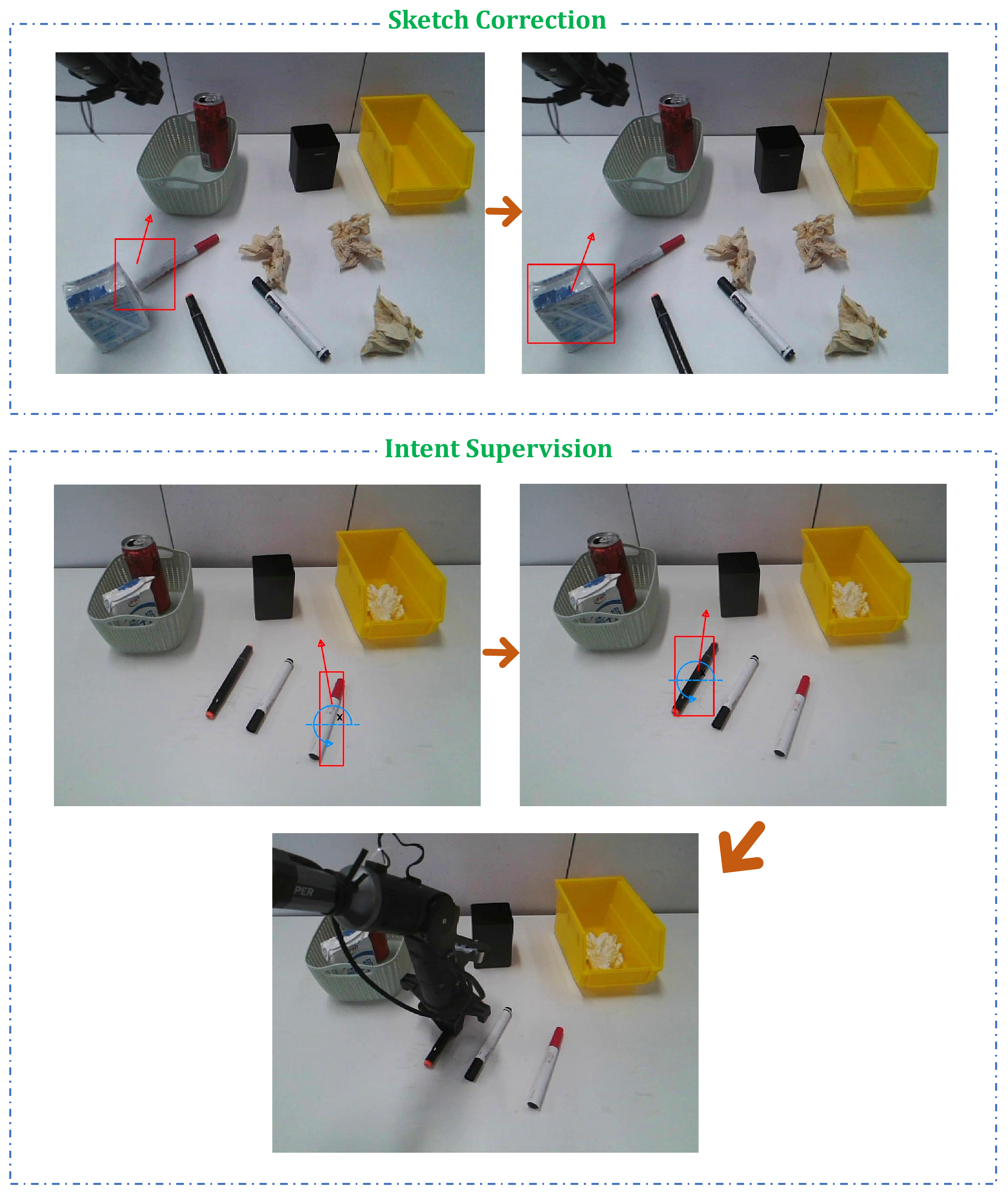}
    \caption{\textbf{Human Interventions via Visual Sketches.}
The explicit intermediate representation serves as an interactive interface.
\textit{(Top) Sketch Correction:} A user identifies a spatially inaccurate sketch (left) and rectifies the bounding box and arrow (right) to prevent execution failure.
\textit{(Bottom) Intent Supervision:} Although the model proposes a valid plan (left), the user modifies the sketch to enforce a specific preference or safety constraint (right), redirecting the robot's behavior in real-time.}
    \label{fig:hum}
\end{figure}

\subsection{Human-in-the-Loop Capabilities}
\label{sec:hitl}

A key advantage of Action-Sketcher's explicit reasoning and sketch generation pipeline is the natural integration of human oversight and intervention. Unlike end-to-end policies that operate as black boxes, our framework's interpretable Visual-Sketches enable humans to inspect, correct, and steer the robot's spatial understanding and intended actions in real-time.

\noindent\textbf{Sketch Correction.} During inference, users can review and refine the model's generated Visual-Sketches after every reasoning step before action execution begins. Fig.~\ref{fig:hum} first subplot illustrates a case where the model's initial Visual-Sketch contains spatial inaccuracies (top) that could lead to task failure. Through our PyQt-based annotation interface, a human operator can quickly identify these errors and provide corrected sketches by adjusting coordinates, repositioning keypoints, or redrawing arrows (bottom). This correction mechanism requires only 3-5 seconds of review per sub-task while ensuring accurate manipulation.

\noindent\textbf{Intent Supervision.} Beyond correcting spatial inaccuracies, our framework enables intent supervision. Even when the model generates spatially accurate Visual-Sketches, human operators may wish to modify the robot's intended behavior to better align with their preferences or safety constraints. Fig.~\ref{fig:hum} second subplot demonstrates how users can steer the model's behavior by modifying the Visual-Sketch after each reasoning step. The model initially generates a valid approach (top), but the human operator redraws the Visual-Sketch to specify an alternative strategy (bottom). This provides transparent and intuitive behavioral control through visual modifications rather than complex policy adjustments.

\section{Future Work}
\label{future}

In this work, we have demonstrated the efficacy of explicit visual grounding in long-horizon manipulation. Looking ahead, we aim to advance Action-Sketcher through five key initiatives:

\begin{itemize}
    \item \textbf{Robust Generalization and Benchmarking.} We will extensively validate the see-think-sketch-act framework's effectiveness across a broader spectrum of simulation benchmarks and real-world scenarios~\cite{nasiriany2024robocasa, zhou2025robotracer, ji2025visualtrans, wang2025towards}, ensuring robust performance in diverse, unstructured environments beyond the in-domain distribution. And we will also further explore using reinforcement fine-tuning (RFT)~\cite{tan2025reason, song2025maniplvm} or policy reinforcement learning (RL)~\cite{amin2025pi, tan2025robo, li2025gr} techniques to improve the model’s generalization.
    
    \item \textbf{Scaling Properties of the Architecture.} We plan to investigate the scaling laws of our approach by employing backbones with significantly larger parameter counts and different model structures~\cite{starvla2025}. Our goal is to assess whether increased model capacity or different model paradigm enhance the spatial precision and semantic fidelity of the visual sketcher's predictions.
    
    \item \textbf{Design Space of Visual Sketches.} We intend to systematically explore how different visual attributes, such as the shape of primitives, color schemes, and line styles, influence policy attention and robustness. This ablation study will identify the optimal visual language for maximizing information transfer between reasoning and acting~\cite{li2025rovi}.
    
    \item \textbf{Expansion to Mobile Manipulation.} Moving beyond static tabletop settings, we aim to extend Action-Sketcher to mobile manipulation tasks~\cite{chen2025ac}. This involves adapting visual sketches to coordinate base mobility with arm manipulation, for instance, by rendering navigation waypoints on the floor or defining long-range interaction targets in large-scale environments.
    
    \item \textbf{Advanced Human-in-the-Loop Paradigms.} Finally, we will explore novel interaction modalities to integrate human feedback with agent system~\cite{tan2025roboos}. We aim to develop intuitive interfaces that allow for real-time sketch correction and intent modification, thereby enhancing decision-making safety in dynamic embodied scenarios.

\end{itemize}

\end{document}